\documentclass[10pt,twocolumn,letterpaper]{article}

\usepackage[pagenumbers]{wacv}              %

\usepackage{newtxtext,newtxmath}

\usepackage{booktabs}
\usepackage{colortbl}
\definecolor{cbest}{HTML}{D6F0DA}
\definecolor{csecond}{HTML}{FFF4C7}
\definecolor{cworst}{HTML}{FAD6D6}
\usepackage{multirow}
\usepackage{graphicx}
\usepackage{bbm}
\usepackage{float}
\usepackage{algorithm}
\usepackage{algpseudocode}

\definecolor{wacvblue}{rgb}{0.21,0.49,0.74}
\usepackage[breaklinks,colorlinks,allcolors=wacvblue]{hyperref}

\def\wacvPaperID{376} %
\def\confName{WACV}
\def\confYear{2027}

\title{\textsc{Sealect}: Scene-Conditioned Pose Priors for Plausible Maritime Vessel Retrieval}

\author{Faisal Mohammad$^{1,2}$ \qquad Mohamed Younis$^{1}$\\
$^{1}$University of Maryland, Baltimore County\\
$^{2}$Johns Hopkins University Applied Physics Laboratory\\
{\tt\small faisalm2@umbc.edu, faisal.mohammad@jhuapl.edu, younis@umbc.edu}
}

\begin{document}
\AddToShipoutPicture*{\AtPageUpperLeft{\put(\LenToUnit{0.5\paperwidth},\LenToUnit{-0.32in}){\makebox(0,0)[t]{\parbox{\textwidth}{\centering\scriptsize Accepted to the IEEE/CVF Winter Conference on Applications of Computer Vision (WACV) 2027.\\ \textcopyright\,2027 IEEE. Personal use of this material is permitted. Permission from IEEE must be obtained for all other uses, in any current or future media, including reprinting/republishing this material for advertising or promotional purposes, creating new collective works, for resale or redistribution to servers or lists, or reuse of any copyrighted component of this work in other works.}}}}}
\maketitle

\begin{abstract}
We address scene-conditioned object retrieval for compositional augmentation in maritime imagery, where strong geometric constraints make appearance-only retrieval produce physically implausible vessel insertions. We propose \textsc{Sealect}, a retrieval framework that jointly models visual compatibility, vessel pose, scale, and scene geometry through a unified score combining neural similarity, scene-conditioned pose priors, and observed-pose consistency. We construct a maritime scene--pose library and model the pose and scale priors nonparametrically, as a kernel that conditions each prior continuously on the scene with no learned parameters. Across seven maritime benchmarks, our trained scene-conditioned retriever exceeds the strongest appearance encoder (SigLIP\,2) in recall ($+0.14$), and the training-free prior raises geometric plausibility further at negligible recall cost, more than halving the constraint violations of appearance retrieval and outperforming learned foreground-search baselines (GALA-style, DiscoFOS). In a blind study of end-to-end composites (BootPlace placement, DiffHarmony harmonization), \textsc{Sealect} ranks first among six methods, ahead of random and GALA-style selection and statistically tied with CLIP, DiscoFOS, and DINO.
\end{abstract}

\section{Introduction}

\begin{figure}[t]
\centering
\includegraphics[width=\columnwidth]{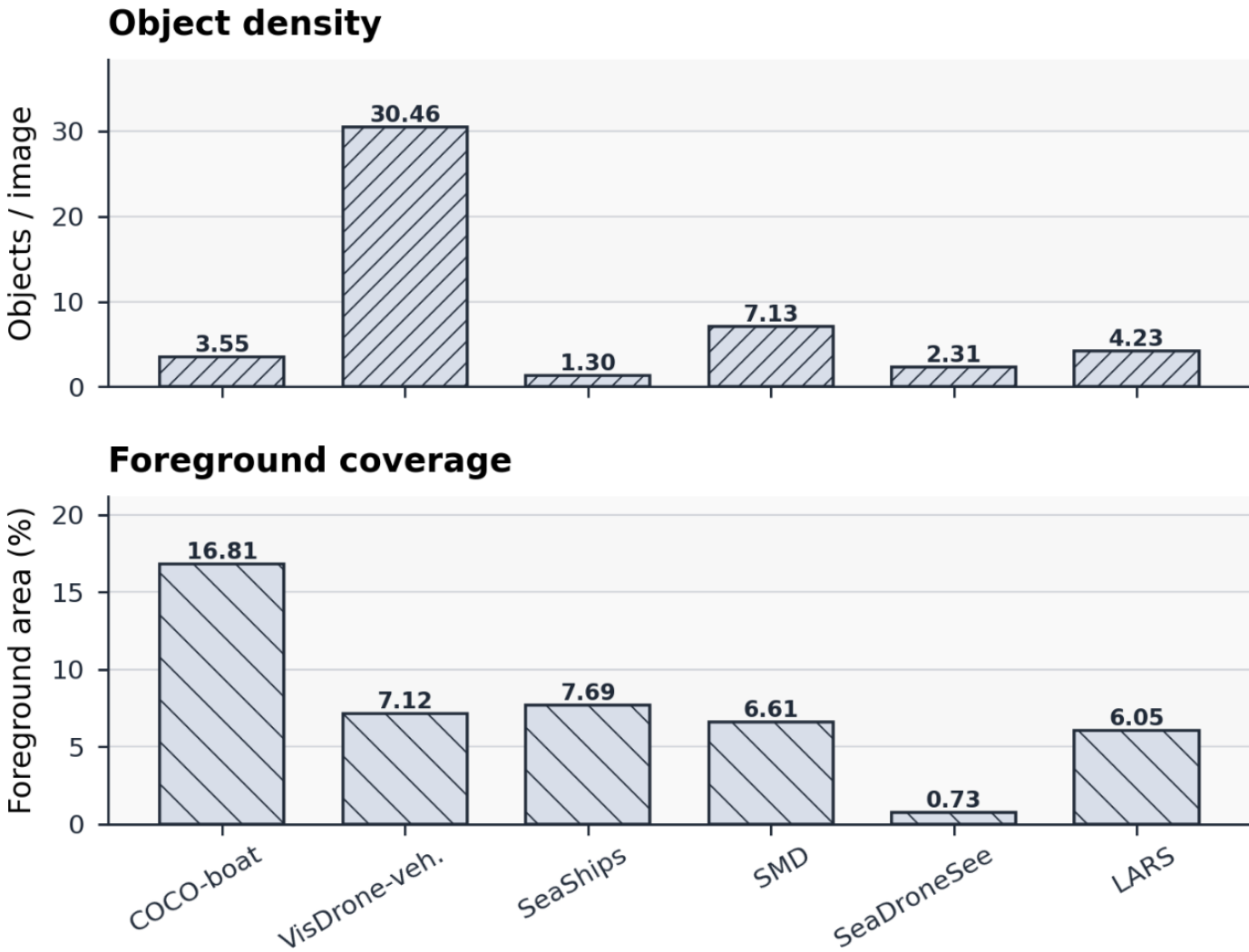}
\caption{Supervision density comparison across some maritime datasets (SeaDronesSee \cite{varga2022seadronessee}, SeaShips \cite{8438999}, SMD \cite{prasad2017video}, LaRS \cite{zust2023larsdiversepanopticmaritime}), boat instances from the general-domain COCO \cite{DBLP:journals/corr/LinMBHPRDZ14} dataset, and vehicle instances from the aerial VisDrone dataset \cite{VISDRONE} (class-positive setting). Maritime datasets exhibit substantially lower object density than aerial data, while all datasets have lower foreground coverage than COCO, highlighting the extreme sparsity of maritime supervision.}
\label{sparsity}
\end{figure}

\setlength\belowcaptionskip{-2ex}
\begin{figure}[t]
\centering
\includegraphics[width=\columnwidth]{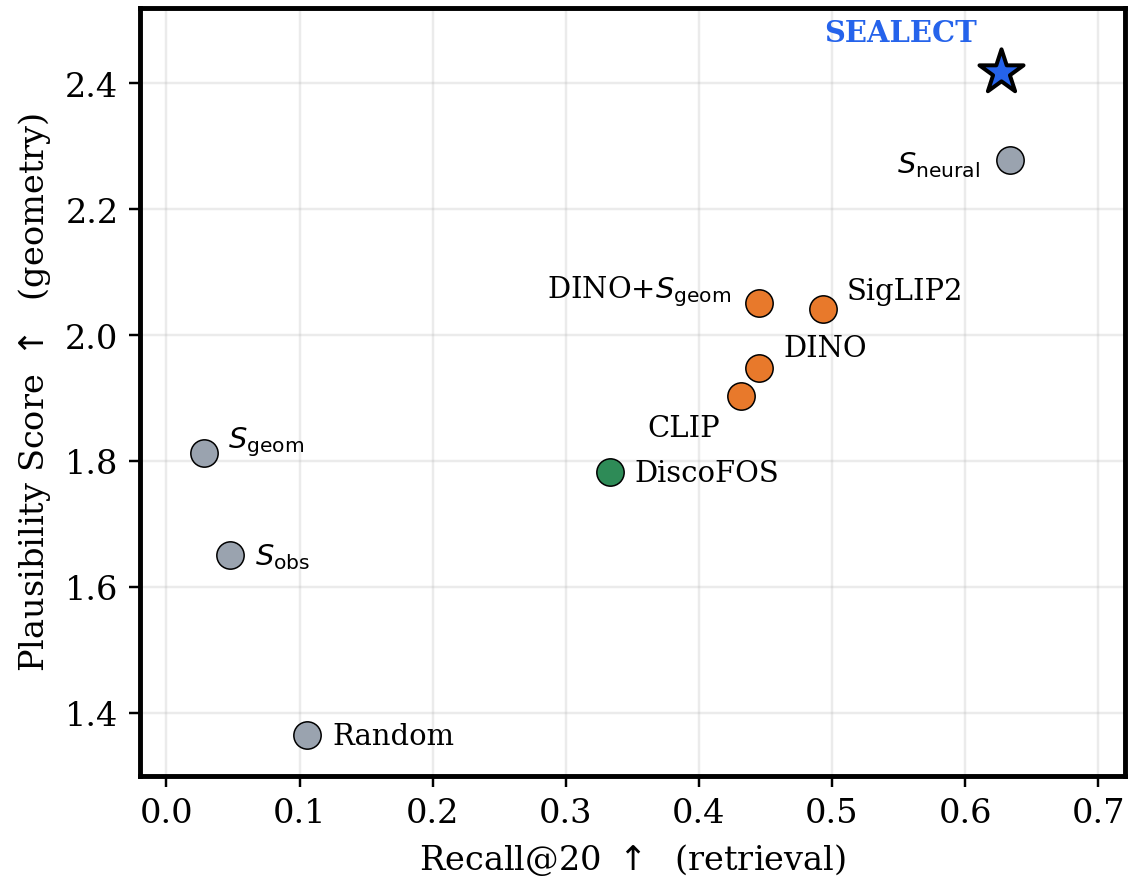}
\caption{\textsc{Sealect} sits at the top-right of the recall--plausibility frontier (Recall@20 vs.\ plausibility score PS on our maritime corpus). Appearance encoders (CLIP, DINO, SigLIP\,2) and DiscoFOS cluster at moderate recall and plausibility; the geometry- and observation-only priors are low-recall ($S_{\text{geom}}$ and DINO$+S_{\text{geom}}$ use the discrete-bin prior). \textsc{Sealect} (star) attains the highest plausibility while nearly matching the best recall.}
\label{fig:pareto}
\end{figure}
Perceiving surface vessels reliably underpins maritime safety and autonomy: collision avoidance for autonomous surface vehicles, vessel-traffic monitoring in ports and coastal waters, search and rescue, and maritime surveillance all depend on detecting and recognizing ships from imagery~\cite{prasad2017video,varga2022seadronessee}. As these systems move toward open-water deployment, they must generalize across the wide range of viewpoints, vessel types, and sea and weather conditions encountered on the water. Surface vessel perception is constrained by limited dataset diversity, leading to poor robustness under distribution shift. While image augmentation is commonly used to address this \cite{xu2023comprehensive, kaur2021data, kumar2024image}, conventional methods (e.g., cropping, color jittering, flipping) ignore scene structure, and poorly chosen transformations can produce out-of-distribution samples that hurt performance \cite{ratner2017learning}. This limitation is especially acute in maritime imagery, where datasets are small and scenes are sparse, dominated by homogeneous water that leaves much of the image uninformative; Figure~\ref{sparsity} contrasts object density and foreground coverage across maritime and general-domain datasets. Increasing the number of object instances per image through copy-paste has been shown to improve detection and instance segmentation \cite{ghiasi2021simplee}. However, naive copy-paste ignores scene context \cite{guo2025cacp}, and in maritime scenes it easily produces geometrically implausible composites (e.g., vessels on land or at an impossible scale). Pasting artifacts can also reduce the benefit of augmentation \cite{dwibedi2017cut}, motivating augmentation strategies that explicitly incorporate geometric and contextual constraints.

Image composition methods \cite{niu2026makingimagesrealagain, zhan2021adversarialimagecompositionauxiliary} aim to generate realistic composites by integrating foreground objects into background scenes. A central component of this problem is \emph{object placement} \cite{CHENG2025131161, zhou2025bootplace, zhu2023topnet, zhou2022learning, zhang2020learning}, which determines plausible locations and scales for inserting objects into a scene. In this work, we do not address placement directly, but instead focus on the complementary problem of \emph{object selection} for maritime compositional augmentation. This setting introduces domain-specific constraints, including water-surface geometry, horizon alignment, and viewpoint-dependent appearance and scale \cite{RuizPonce2023POSEIDON}, which make naive selection of candidate objects prone to unrealistic configurations. Compounding this, large-scale annotated datasets for maritime object placement and selection remain limited. While the Object Placement Assessment (OPA) dataset \cite{liu2021OPA} provides a benchmark for evaluating placement plausibility, it does not explicitly model the relationship between object pose and scene context.

We observe that vessel pose and environmental context exhibit a structured dependency. In maritime imagery, plausible vessel configurations are tightly coupled to scene geometry: near-nadir viewpoints produce water-dominated scenes with top-down vessel appearances, whereas side-view perspectives yield structured sky--vessel--water layouts. This regularity implies that scene context strongly constrains plausible vessel configurations. Motivated by this observation, we formulate vessel augmentation as a scene-conditioned retrieval problem. Given a background image \(x_{\text{bg}}\), we extract a global scene representation \(\mathbf{c}=g(x_{\text{bg}})\) together with pose-aware geometric cues derived from the scene. Candidate vessel crops \(o_j \in \mathcal{O}\) are embedded into a shared retrieval space as \(e_j=f_{\text{obj}}(o_j)\), while the background scene is encoded as a scene-conditioned query $q = f_{\text{scene}}(\mathbf{c})$. Candidates are ranked using a compatibility score that combines learned visual similarity with scene-conditioned geometric and observation-consistency constraints,
\begin{equation}
S =
\lambda_n S_{\text{neural}}
+
\lambda_g S_{\text{geom}}
+
\lambda_o S_{\text{obs}}.
\end{equation}

Top-ranked candidates are subsequently composited into the background using off-the-shelf placement and harmonization models, producing maritime augmentations. We study the \emph{selection} step and evaluate \textsc{Sealect} directly as a retrieval method, measuring ranking, distributional alignment, and plausibility (Figure~\ref{fig:pareto}).

\smallskip
\noindent\textbf{Contributions.} (i) We recast maritime vessel augmentation as \emph{scene-conditioned selection}, which vessel to insert rather than where to place it or the nearest appearance match, exploiting that a vessel's plausible pose, scale, and placement are strongly constrained by the background viewpoint and water/sky/horizon layout. (ii) We realize this in \textsc{Sealect}, which fuses a learned retriever with a \emph{training-free, nonparametric} prior that reads a vessel's plausible geometry off a compact scene descriptor. Because maritime pose supervision is too sparse to learn such a prior reliably, this explicit formulation outperforms learned discriminative compatibility (GALA-style, DiscoFOS) on retrieval and plausibility metrics, and in a blind study of end-to-end composites \textsc{Sealect} ranks first while remaining statistically tied with CLIP, DINO, and DiscoFOS.

\section{Related Work}
\begin{figure*}[t]
    \centering
    \includegraphics[width=\textwidth]{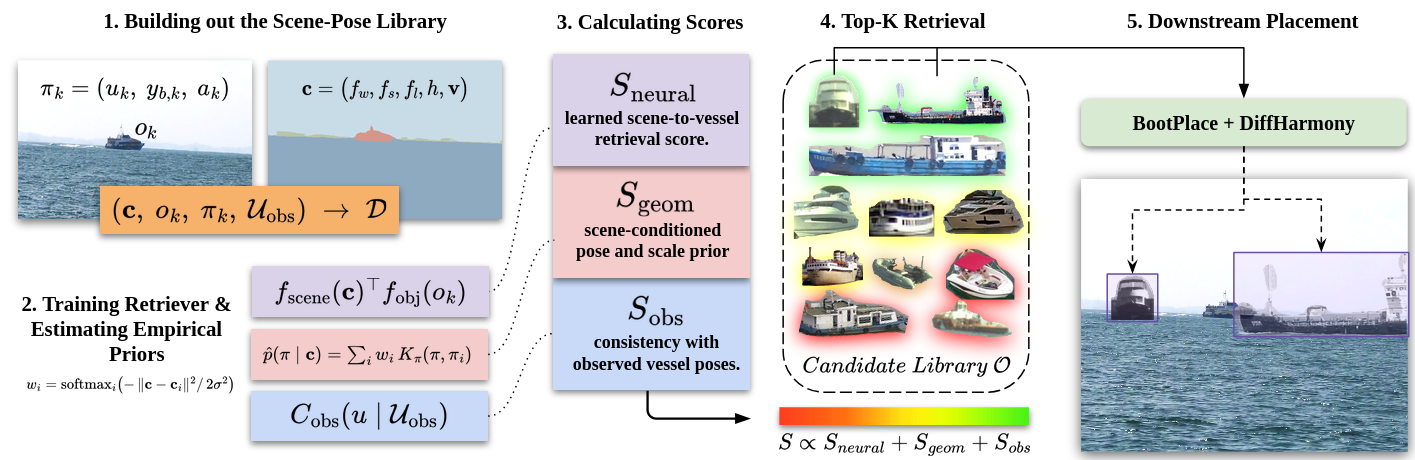}
    \caption{Overview of the proposed \textsc{Sealect} pipeline. A scene--pose library is first constructed from maritime images by extracting scene descriptors and vessel pose metadata. Empirical scene-conditioned priors are then estimated and combined with neural retrieval and observed-pose consistency scores to rank candidate vessel instances. Top-ranked candidates are subsequently passed to BootPlace for downstream placement and compositional augmentation.}
    \label{fig:pipeline}
\end{figure*}
\noindent\textbf{Compositional Augmentation \& Object Retrieval.}
Copy--paste augmentation raises object density by compositing segmented instances into new scenes \cite{dwibedi2017cut,ghiasi2021simplee} or by jittering existing instances \cite{fang2019instaboost}. Meanwhile, retrieval-based augmentation \emph{selects} real instances via strong encoders (SimCLR \cite{chen2020simple}, MoCo \cite{he2020momentum}, BYOL \cite{grill2020byol}, CLIP \cite{radford2021learning}, ALIGN \cite{jia2021align}, BLIP \cite{li2022blip}, DINO/DINOv2 \cite{caron2021emerging,oquab2024dinov2}, SigLIP\,2 \cite{tschannen2025siglip2}) or composed retrieval driven by an auxiliary query \cite{vo2019tirg,baldrati2022effective,liu2021cirr,saito2023pic2word} and place or instance retrieval \cite{arandjelovic2016netvlad,noh2017delf,radenovic2018finetuning}. Appearance similarity alone ignores the geometric consistency that compositional realism demands \cite{lin2018stgan,liu2021OPA,niu2022fopa}. The closest task, \emph{foreground object search} (FOS), learns compatibility embeddings \cite{zhao2018compositing,zhao2019ufo}, adds geometry-and-lighting awareness (GALA \cite{gala}), or distills a composite-image feature into a two-encoder retriever (DiscoFOS \cite{discofos}). Recent work shifts to \emph{generative} insertion \cite{winter2025objectmate} and real composition datasets \cite{kim2025orida}. None of these selects candidates with an explicit, pose-conditioned scene prior: GALA learns geometric compatibility discriminatively, and the maritime tool POSEIDON~\cite{RuizPonce2023POSEIDON} relies on capture metadata. Selecting real instances also avoids the synthetic-to-real gap of generated insertions.

\smallskip\noindent\textbf{Object Placement.}
The complementary problem of \emph{where} to insert forms our pipeline's final stage: placements are synthesized adversarially with geometric/color consistency (TERSE \cite{tripathi2019learning}, GCC-GAN \cite{chen2019toward}), learned as distributions via graph completion or object-removal self-supervision (GracoNet \cite{zhou2022learning}, PlaceNet \cite{zhang2020learning}), predicted as location--scale heatmaps (TopNet \cite{zhu2023topnet}), or cast as detection on object-masked scenes (BootPlace \cite{zhou2025bootplace}). Unlike PlaceNet, which uses the boxes of removed objects as placement supervision, we use boxes to extract scene-aligned vessel orientations, and we train BootPlace on maritime imagery for downstream evaluation.

\smallskip\noindent\textbf{Pose Estimation.}
Beyond human pose estimation \cite{pfister2015flowingconvnetshumanpose,9157713}, orientation work models angular uncertainty with directional distributions \cite{prokudin2018deep,mohlin2020probabilistic} or regresses a continuous 6D rotation on $SO(3)$ \cite{zhou2020continuityrotationrepresentationsneural}; others detect 3D boxes with orientation \cite{10658487,Xiong_2023,li2024unimode}. In the maritime domain, \cite{kiefer2023stableyawestimationboats} targets probabilistic yaw with temporal aggregation and \cite{holst2025fusingmonocularrgbimages} derives vessel pose labels (BONK-pose) by fusing RGB with AIS, assuming zero roll and pitch; we convert these labels to augment our own.

\smallskip\noindent\textbf{Maritime Benchmarks.}
Existing datasets support detection, tracking, and segmentation under diverse conditions: SeaShips \cite{8438999}, SMD \cite{prasad2017video}, LaRS \cite{zust2023larsdiversepanopticmaritime}, ABOShips \cite{iancu2021aboshipsinshoreoffshore}, and KOLOMVERSE \cite{10689331} capture on-board/on-shore views, while UAV-based SeaDronesSee \cite{varga2022seadronessee} adds distinct viewing geometries and scales. SeaSAW \cite{Kaur_2022_CVPR} ($\sim$1.9M images) is not publicly available, and MARVEL \cite{MARVEL} offers fine-grained categories but mostly close-range web photographs intended for classification. None provide consistent, large-scale vessel-pose annotations for learning pose-conditioned object--scene relationships; where partial pose metadata exists \cite{articleremsense}, it is not designed for this purpose.

\section{Methodology: \textsc{Sealect}}

The \textsc{Sealect} pipeline consists of three stages (Figure~\ref{fig:pipeline}): (1) curating an appearance-aligned maritime pose dataset and training a monocular pose estimator, (2) deriving scene-conditioned pose and scale priors, and (3) integrating these components into a retrieval framework for selecting geometrically plausible vessel candidates for downstream placement.

\subsection{Appearance-Aligned Pose Estimation}

We construct an appearance-aligned maritime pose dataset by annotating vessel orientation in a Unity3D interface using a canonical 3D vessel model. Annotations capture the \emph{apparent pose} of the vessel in image space rather than its true world-frame orientation, which is more appropriate for compositional retrieval and placement. Our training set combines 947 manually annotated instances across multiple maritime benchmarks, their horizontal flips, and 3,753 converted BONK-Pose samples \cite{holst2025fusingmonocularrgbimages}, spanning aerial, on-board, and shoreline viewpoints. Given a vessel crop $I$, we estimate orientation $\hat{R}\in SO(3)$ with a dual-stream network that fuses CNN geometric features with a SegFormer semantic branch pretrained for vessel segmentation, predicting a continuous 6D rotation \cite{zhou2020continuityrotationrepresentationsneural} together with discretized pose bins. It is trained with geodesic and bin-classification losses under a flipped-heading hypothesis that models the $180^\circ$ bow/stern symmetry (full architecture and losses in supplementary Section~\ref{sec:pose_supp}). An earlier run of the same architecture, trained on 804 manual annotations only, reaches $5.9^\circ/7.7^\circ$ pitch/roll MAE and a symmetry-aware geodesic error of $23.0^\circ$ mean on a 200-vessel validation set from eight datasets that was also used for checkpoint selection; yaw is resolved only up to the $180^\circ$ bow/stern ambiguity (Table~\ref{tab:pose_error}).

\subsection{Scene--Pose Library Construction}

We build a scene--pose library capturing the joint statistics of background context and vessel geometry. From a grounded segmentation~\cite{liu2024groundingdino,ravi2024sam2} of each image (supplementary Section~\ref{sec:impl}), with annotated vessel regions masked out so it reflects the background alone, we form a scene descriptor
\begin{equation}
\mathbf{c} = \big( f_w, f_s, f_l, h, \mathbf{v} \big),
\end{equation}
of water, sky, and land fractions $f_w, f_s, f_l \in [0,1]$, normalized horizon location $h \in [0,1]$, and a vertical band profile $\mathbf{v} \in \mathbb{R}^B$; a coarse context key $\gamma=\kappa(\mathbf{c})$, which hard-quantizes the viewpoint and layout classes, the three fractions, and $h$, supports the discrete-bin ablation of Section~\ref{sec:experiments}. For each annotated vessel we record the pose bins $u=(b_p, b_y, b_r)$ predicted by the estimator's bin classifiers (e.g., nadir vs.\ level pitch, frontal vs.\ side-view yaw) that stabilize statistics under noisy predictions, together with its vertical-placement bin $y_b$ and relative scale $a$, giving the joint geometric state
\begin{equation}
\pi = (u,\, y_b,\, a).
\end{equation}
We additionally collect the poses $\mathcal{U}_{\text{obs}}$ of the other vessels in each scene, which capture local regularities the global descriptor misses. Over all $N$ annotated vessels (with crops $o_i$), the library $\mathcal{D}=\{(\mathbf{c}_i, o_i, \pi_i, \mathcal{U}_{\text{obs},i})\}_{i=1}^{N}$ defines the empirical distribution from which we estimate the priors below.

\begin{table*}[t]
\centering
\small
\setlength{\tabcolsep}{8pt}
\renewcommand{\arraystretch}{1.15}
\begin{tabular}{l ccc|ccc}
\toprule
& \multicolumn{3}{c|}{$K=20$} & \multicolumn{3}{c}{$K=100$} \\
\textbf{Methods} & Recall $\uparrow$ & NDCG $\uparrow$ & PS $\uparrow$ & Recall $\uparrow$ & NDCG $\uparrow$ & PS $\uparrow$ \\
\midrule
Random & 0.106 & \cellcolor{cworst}0.726 & \cellcolor{cworst}1.364 & 0.347 & \cellcolor{cworst}0.719 & \cellcolor{cworst}1.365 \\
CLIP & 0.432 & 0.801 & 1.903 & 0.645 & 0.782 & 1.731 \\
DINO & 0.445 & 0.810 & 1.947 & 0.637 & 0.791 & 1.776 \\
SigLIP\,2~\cite{tschannen2025siglip2} & 0.493 & 0.814 & 2.042 & 0.671 & 0.799 & 1.843 \\
DiscoFOS~\cite{discofos} (zero-shot) & 0.071 & 0.786 & 1.414 & 0.164 & 0.789 & 1.459 \\
DiscoFOS~\cite{discofos} (fine-tuned) & 0.333 & 0.770 & 1.783 & 0.575 & 0.758 & 1.730 \\
GALA-style~\cite{gala} (geom.) & 0.107 & 0.813 & 1.672 & 0.194 & 0.812 & 1.695 \\
DINO $+$ $S_{\text{geom}}^\dagger$ & 0.445 & 0.822 & 2.051 & 0.611 & 0.809 & 1.907 \\
\midrule
$\lambda_n{=}1,\lambda_g{=}0,\lambda_o{=}0$ ($S_{\text{neural}}$ only) & \cellcolor{cbest}\textbf{0.634} & 0.830 & 2.278 & \cellcolor{cbest}\textbf{0.809} & 0.814 & 2.066 \\
$\lambda_n{=}0,\lambda_g{=}1,\lambda_o{=}0$ ($S_{\text{geom}}$ only)$^\dagger$ & \cellcolor{cworst}0.028 & \cellcolor{cbest}\textbf{0.940} & 1.812 & \cellcolor{cworst}0.039 & \cellcolor{cbest}\textbf{0.935} & 1.825 \\
$\lambda_n{=}0,\lambda_g{=}0,\lambda_o{=}1$ ($S_{\text{obs}}$ only) & 0.048 & \cellcolor{csecond}0.853 & 1.651 & 0.078 & \cellcolor{csecond}0.855 & 1.671 \\
\midrule
\textsc{Sealect} (discrete) & 0.598 & 0.843 & \cellcolor{csecond}2.354 & 0.767 & 0.827 & \cellcolor{csecond}2.142 \\
\textsc{Sealect} (kernel) & \cellcolor{csecond}0.627 & 0.848 & \cellcolor{cbest}\textbf{2.418} & \cellcolor{csecond}0.801 & 0.832 & \cellcolor{cbest}\textbf{2.212} \\
\bottomrule
\end{tabular}
\caption{Retrieval performance for maritime surface vessel candidate selection under constrained ($K{=}20$) and expanded ($K{=}100$) candidate pools. \textsc{Sealect} uses the kernel-conditioned scene prior (Eq.~\ref{eq:kernel_prior}) with fixed weights $(\lambda_n,\lambda_g,\lambda_o){=}(1,1,0.10)$. Component-only rows ($S_{\text{neural}}$/$S_{\text{geom}}$/$S_{\text{obs}}$ only) ablate the fusion. Shading marks the best (green), second-best (yellow), and worst (red) value per column in all result tables (ties share a shade). PS is the mean number (0--4) of geometric attributes (pitch, yaw, placement, scale) on which a retrieved candidate matches the held-out vessel; NDCG orders by the same relevance. $^\dagger$Uses the discrete-bin prior. CLIP is ViT-B/32, DINO is DINOv2 ViT-S/14, and SigLIP\,2 is B/16; all three rank by similarity to the query vessel's crop.}
\label{tab:retrieval_combined}
\end{table*}

\subsection{Scene-Conditioned Pose Priors}

Given $\mathcal{D}$, we estimate a scene-conditioned prior over the joint geometric state $\pi=(u,y_b,a)$, comprising discretized pose $u$, vertical-placement bin $y_b$, and relative scale $a$, as a scene-weighted empirical average over the library,
\begin{equation}
\hat p(\pi \mid \mathbf{c}_q)
=
\sum_i w_i(\mathbf{c}_q)\, k_\pi(\pi, \pi_i),
\label{eq:kernel_prior}
\end{equation}
where $k_\pi$ is a kernel over the discretized geometric state and the scene weights $w_i(\mathbf{c}_q)$, normalized over the library, set how much each entry contributes to the query $\mathbf{c}_q$. Written this way, the prior exposes a single design choice: the scene weighting. A discrete context bin $\gamma=\kappa(\mathbf{c})$ is the hard-indicator special case $w_i(\mathbf{c}_q)\propto\mathbf{1}[\kappa(\mathbf{c}_i){=}\kappa(\mathbf{c}_q)]$, averaging only over scenes that share a quantized bin. We instead weight every scene smoothly with an RBF over standardized descriptors,
\begin{equation}
w_i(\mathbf{c}_q)
=
\operatorname{softmax}_i\!\big(-\lVert \mathbf{c}_q - \mathbf{c}_i \rVert^2 / 2\sigma^2\big),
\label{eq:kernel_weights}
\end{equation}
of bandwidth $\sigma$, a Nadaraya--Watson-type estimate~\cite{nadaraya1964estimating,watson1964smooth}: as $\sigma\!\to\!0$ the weights collapse onto the nearest library entries, and as $\sigma\!\to\!\infty$ they approach the global marginal. The kernel acts on a standardized copy of the vector that feeds the scene tower, $\mathbf{c}$ augmented with the target region's position, size, and aspect and with viewpoint/layout descriptors (supplementary Section~\ref{sec:impl}), so an entry is up-weighted when both its background and its vessel's region resemble the query's target (median ${\approx}15$ effective entries at $\sigma{=}1$; supplementary Section~\ref{sec:extra}). Both priors are fully empirical; since a hard bin must coarsen the scene to stay populated, $\kappa$ quantizes each quantity into a few levels, whereas the kernel conditions on the full continuous descriptor without empty bins. The prior factorizes as
\begin{equation}
\hat p(\pi \mid \mathbf{c}_q)
\propto
\hat p(u \mid \mathbf{c}_q)\,
\hat p(y_b \mid \mathbf{c}_q)\,
\hat p(a \mid \mathbf{c}_q),
\label{eq:factorized_prior}
\end{equation}
treating pose, placement, and scale as conditionally independent given the target-aware descriptor, so each factor is a scene-weighted statistic read off a single axis: the pose marginals $\hat p(b_p\mid\mathbf{c}_q),\hat p(b_y\mid\mathbf{c}_q),\hat p(b_r\mid\mathbf{c}_q)$, the vertical placement $\hat p(y_b\mid\mathbf{c}_q)$, and a Gaussian over log-scale with kernel-weighted mean and variance. (The discrete variant, whose key lacks target geometry, instead uses $\hat p(y_b\mid u,\gamma)$ and $\hat p(a\mid y_b,\gamma)$.) A candidate's own pose, placement, and scale are scored under this target-conditioned distribution. These capture structured maritime regularities such as horizon-aligned distant vessels and larger near-field vessels lower in the frame; the factor weights emphasize pitch (the reliably estimated axis and the dominant cue separating nadir from level viewpoints) over the more ambiguous yaw. While $\hat p(\pi \mid \mathbf{c}_q)$ captures global scene statistics, individual scenes often exhibit additional structure. To account for this, we define an observation-conditioned compatibility term over discretized poses
\begin{equation}
C_{\text{obs}}(u \mid \mathcal{U}_{\text{obs}})
=
(1-\alpha)U(u)
+
\alpha
\frac{1}{|\mathcal{U}_{\text{obs}}|}
\sum_{u_i \in \mathcal{U}_{\text{obs}}}
k_u(u,u_i),
\label{eq:obs_compat}
\end{equation}
where $U(u)$ denotes a uniform baseline over discretized pose bins and $k_u(\cdot,\cdot)$ is a kernel measuring local similarity in pose-bin space. This preserves multimodal support while favoring configurations consistent with observed scene geometry.

\subsection{Scene-Conditioned Retrieval}

Given a query scene $\mathbf{c}$ and a target region to fill, we retrieve candidate vessels $\{o_j\}$ that are both geometrically plausible and visually compatible. Scene and candidate instances are embedded as
\begin{equation}
q = f_{\text{scene}}(\mathbf{c}), \quad e_j = f_{\text{obj}}(o_j),\end{equation}
where embeddings are $\ell_2$-normalized and the scene query encodes both the background statistics and the target region's local placement geometry (supplementary Section~\ref{sec:impl}). The neural compatibility score is defined as
\begin{equation}
S_{\text{neural}}(j) = q^\top e_j,
\end{equation}
a learned scene--candidate compatibility over appearance (DINOv2 ViT-S/14~\cite{oquab2024dinov2} and pose-estimator crop features) and geometric context (supplementary Section~\ref{sec:impl}); the explicit prior $S_{\text{geom}}$ adds complementary, training-free geometric structure. The encoders are trained using in-batch contrastive learning~\cite{oord2018cpc,radford2021learning}, which scores each scene query's own vessel above the other candidates in the batch. At inference time, retrieval is formulated using the following scores:
\begin{equation}
S_j =
\lambda_n S_{\text{neural}}(j)
+
\lambda_g S_{\text{geom}}(j)
+
\lambda_o S_{\text{obs}}(j)
\end{equation}
\vspace{0.5pt}
\begin{equation}
S_{\text{geom}}(j)
=
\log \hat p(\pi_j \mid \mathbf{c}_q).
\label{eq:context_score}
\end{equation}
\vspace{0.5pt}
\begin{equation}
S_{\text{obs}}(j)
=
\log C_{\text{obs}}(u_j\mid\mathcal{U}_{\text{obs}}).
\end{equation}

where \(\pi_j=(u_j,y_{b,j},a_j)\); their sum is a hybrid compatibility score, not a normalized probability.

\section{Experiments and Results}
\label{sec:experiments}

We evaluate retrieval along three axes: ranking (candidate relevance under geometric constraints), distributional alignment (match to scene-conditioned statistics without mode collapse), and plausibility (likelihood and constraint violations). Metric definitions are in supplementary Section~\ref{sec:eval_metrics}, and a rendered-pipeline evaluation is in supplementary Section~\ref{sec:rendered}.

\subsection{Retrieval Quality}

We evaluate retrieval on a unified maritime corpus spanning seven benchmarks (six public, plus one access-restricted in-house UAV set\footnote{Provided by the Johns Hopkins University Applied Physics Laboratory.}; supplementary Section~\ref{sec:impl}) under constrained ($K{=}20$) and expanded ($K{=}100$) candidate pools. Appearance-based retrieval, even the strongest encoder SigLIP\,2 (recall $0.493$, PS $2.042$ at $K{=}20$), frequently retrieves vessels with incompatible pose and scale (NDCG $\approx0.80$, PS $<2.05$). The scene prior $S_{\text{geom}}$ attains the highest ranking quality (NDCG $0.940$/$0.935$) but very low recall, showing that geometric statistics alone are insufficient for instance-level retrieval, whereas pure neural retrieval $S_{\text{neural}}$ achieves the highest recall ($0.634$/$0.809$) but is the least geometrically constrained.

The recall advantage over appearance encoders thus comes from the trained retriever ($S_{\text{neural}}$ alone exceeds SigLIP\,2 by $+0.14$ Recall@20), not from the prior: adding $S_{\text{geom}}$ costs a little recall ($0.634{\rightarrow}0.627$ at $K{=}20$, $0.809{\rightarrow}0.801$ at $K{=}100$) and improves ranking quality (NDCG $0.830{\rightarrow}0.848$) and plausibility (PS $2.278{\rightarrow}2.418$) at $K{=}20$. The full model exceeds every appearance and learned-FOS baseline on all three metrics; only the $S_{\text{geom}}$- and $S_{\text{obs}}$-only ablations rank higher on NDCG, at $0.028$ and $0.048$ recall. Reverting the kernel (Eq.~\ref{eq:kernel_prior}) to a discrete context bin lowers all six metrics (\eg, recall $0.627{\rightarrow}0.598$ at $K{=}20$); the kernel keeps its NDCG and PS advantage at every tested bandwidth (recall only for $\sigma{\le}1$) and raises Recall@20 on six of seven datasets.

\subsection{Distributional Alignment}

\begin{figure}[t]
\centering
\includegraphics[width=\columnwidth]{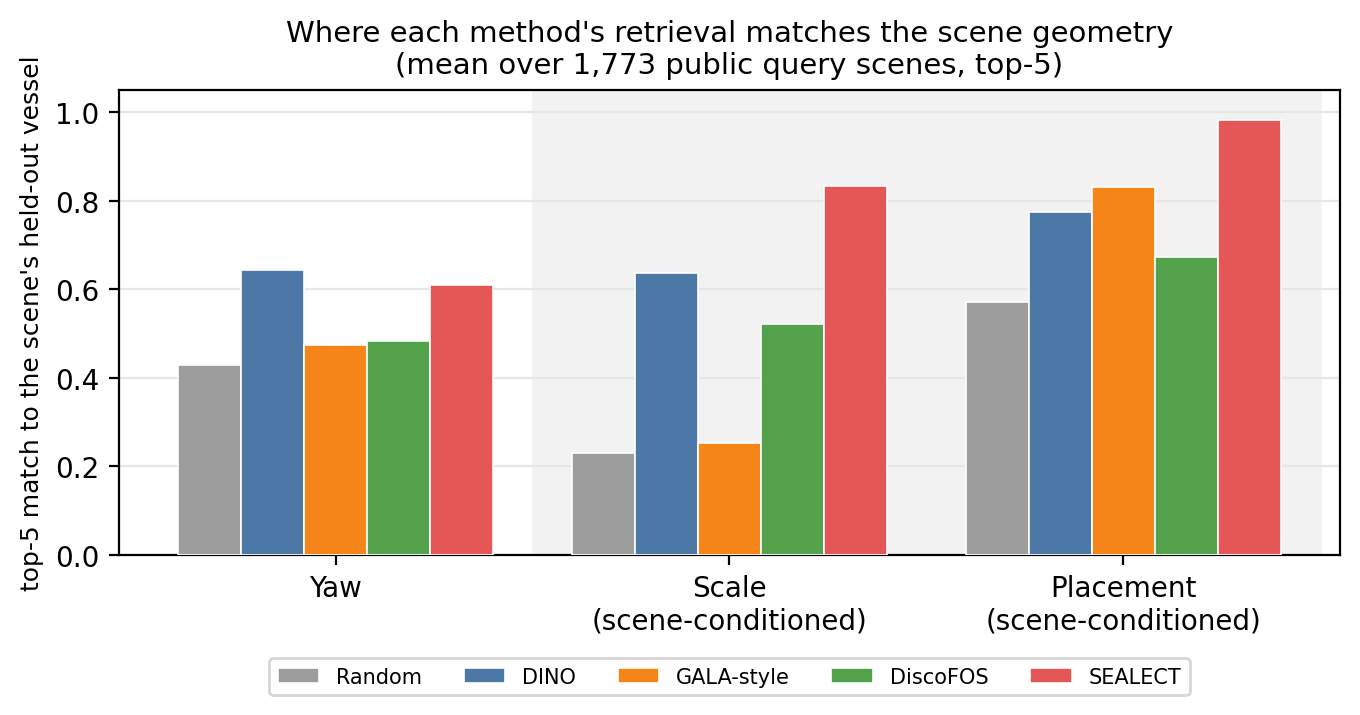}
\caption{Per-axis geometric quality of each method's ranking over $1{,}773$ public query scenes (ABOShips excluded for unreliable metadata; restricted candidates skipped): the fraction of top-5 retrievals matching the scene's held-out vessel in yaw (within one $20^\circ$ bin), scale (within $2\times$), and placement (within one vertical bin). Because \textsc{Sealect} conditions on the target region, it retrieves vessels whose scale and placement fit the scene far more often than appearance (DINO) or learned foreground-search (GALA-style, DiscoFOS) retrieval; on yaw, which is appearance-driven and subject to the $180^\circ$ bow/stern ambiguity, DINO and \textsc{Sealect} lead.}
\label{fig:axis_decomp}
\end{figure}

\begin{figure*}[t]
    \centering
    \includegraphics[width=\textwidth]{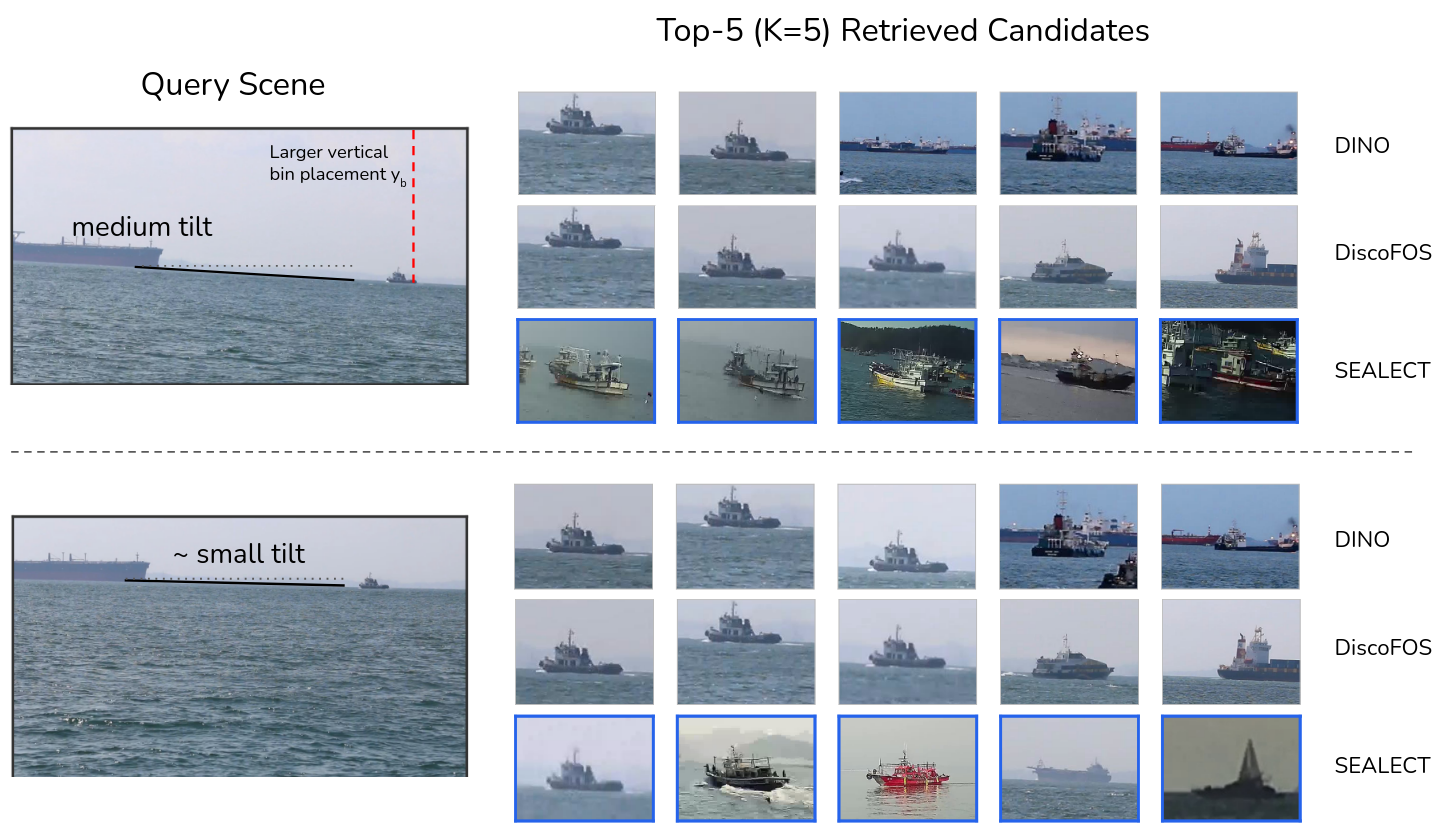}
    \caption{\textsc{Sealect} conditions retrieval on scene viewpoint; appearance and learned-search retrieval do not. Two query scenes hold the \emph{same} target tug but differ in camera viewpoint and placement bin $y_b$ (near-field, top; near the horizon, bottom). \textsc{Sealect}'s top-$5$ shares $0/5$ candidates between them, shifting to higher-pitch near-field vessels (top) and flat, side-facing vessels (bottom), whereas DINO ($4/5$ shared) and DiscoFOS ($5/5$) return nearly the same candidates.}
    \label{fig:viewpoint}
\end{figure*}

We next assess whether retrieved candidates follow the scene-conditioned distributions without collapsing retrieval diversity (Table~\ref{tab:dist_compact}; full per-axis results in supplementary Table~\ref{tab:dist_exp2}). Appearance-based retrieval (CLIP, DINO) attains low Joint-Pose JS but high entropy ($H_{\text{ret}}\approx3.3$), sampling broadly across configurations, whereas prior-only retrieval collapses to a few dominant modes (entropy $0.12$, Joint-Pose Gap $2.66$). \textsc{Sealect} sits between the two: relative to $S_{\text{neural}}$ it reduces Joint-Pose entropy ($3.28{\rightarrow}2.40$ at $K{=}20$) and improves Joint-Pose alignment (Gap $0.68{\rightarrow}0.42$ at $K{=}100$), correcting neural over-dispersion without collapsing diversity. Kernel conditioning also tightens placement alignment over the discrete bin (Y-bin JS $0.0668{\rightarrow}0.0515$ at $K{=}20$). Per axis (Figure~\ref{fig:axis_decomp}), \textsc{Sealect}'s top-5 matches the scene's scale and placement far more often than DINO, GALA-style, or DiscoFOS, and on yaw it trails only DINO; on a controlled pair (Figure~\ref{fig:viewpoint}), \textsc{Sealect}'s retrievals change with viewpoint while those of the baselines stay nearly fixed.

\begin{table}[t]
\centering
\small
\setlength{\tabcolsep}{4pt}
\resizebox{\columnwidth}{!}{
\begin{tabular}{l ccc|ccc}
\toprule
& \multicolumn{3}{c|}{$K=20$} & \multicolumn{3}{c}{$K=100$} \\
\textbf{Method} & JS$\downarrow$ & Gap$\downarrow$ & $H_{\text{ret}}$ & JS$\downarrow$ & Gap$\downarrow$ & $H_{\text{ret}}$ \\
\midrule
Random & 0.241 & 0.969 & 3.72 & 0.238 & 1.016 & 3.78 \\
DINO & \cellcolor{cbest}\textbf{0.159} & \cellcolor{csecond}0.531 & 3.29 & \cellcolor{cbest}\textbf{0.166} & 0.639 & 3.41 \\
DiscoFOS (ft) & 0.196 & 0.534 & 3.21 & 0.186 & 0.616 & 3.37 \\
GALA-style & 0.444 & 0.923 & 1.99 & 0.354 & 0.667 & 2.39 \\
$S_{\text{geom}}$ only$^\dagger$ & \cellcolor{cworst}0.548 & \cellcolor{cworst}2.659 & 0.12 & \cellcolor{cworst}0.539 & \cellcolor{cworst}2.625 & 0.15 \\
$S_{\text{neural}}$ only & \cellcolor{csecond}0.185 & 0.548 & 3.28 & 0.186 & 0.683 & 3.43 \\
\midrule
\textsc{Sealect} (discrete) & 0.198 & 0.559 & 2.46 & 0.191 & \cellcolor{csecond}0.499 & 2.64 \\
\textsc{Sealect} (kernel) & \cellcolor{csecond}0.185 & \cellcolor{cbest}\textbf{0.525} & 2.40 & \cellcolor{csecond}0.178 & \cellcolor{cbest}\textbf{0.415} & 2.59 \\
\bottomrule
\end{tabular}
}
\caption{Distributional alignment (Joint-Pose). \textsc{Sealect} corrects neural over-dispersion (lower retrieved entropy $H_{\text{ret}}$ than $S_{\text{neural}}$) without the prior-only collapse ($H_{\text{ret}}{=}0.12$), achieving the lowest Joint-Pose Gap. Full per-axis (yaw/placement/scale) results in supplementary Table~\ref{tab:dist_exp2}. $H_{\text{ret}}$ has no preferred direction and is unshaded. $^\dagger$Discrete-bin prior.}
\label{tab:dist_compact}
\end{table}

\subsection{Plausibility Evaluation}

Plausibility reflects how often retrieved candidates violate scene-conditioned pose and placement constraints (Table~\ref{tab:plaus_compact}; full per-axis in supplementary Table~\ref{tab:plausibility_combined}). Appearance-based retrieval shows high violation rates despite moderate likelihoods (DINO Joint violation $6.6\%$/$7.4\%$); scene priors sharply reduce these (DINO+$S_{\text{geom}}$ $2.1\%$/$2.3\%$), and prior-only attains the lowest rates ($0.6\%$/$0.6\%$) but at the cost of diversity and instance-level coverage.

\begin{table}[t]
\centering
\small
\setlength{\tabcolsep}{5pt}
\resizebox{\columnwidth}{!}{
\begin{tabular}{l cc|cc}
\toprule
& \multicolumn{2}{c|}{$K=20$} & \multicolumn{2}{c}{$K=100$} \\
\textbf{Method} & LL$\uparrow$ & Viol.$\downarrow$ & LL$\uparrow$ & Viol.$\downarrow$ \\
\midrule
Random & \cellcolor{cworst}-3.053 & \cellcolor{cworst}.120 & \cellcolor{cworst}-3.054 & \cellcolor{cworst}.119 \\
DINO & -2.699 & .066 & -2.751 & .074 \\
DiscoFOS (ft) & -2.790 & .075 & -2.806 & .077 \\
GALA-style & -2.861 & .083 & -2.817 & .081 \\
$S_{\text{geom}}$ only$^\dagger$ & \cellcolor{cbest}\textbf{-1.963} & \cellcolor{cbest}\textbf{.006} & \cellcolor{cbest}\textbf{-1.968} & \cellcolor{cbest}\textbf{.006} \\
$S_{\text{neural}}$ only & -2.742 & .066 & -2.787 & .069 \\
\midrule
\textsc{Sealect} (discrete) & -2.354 & \cellcolor{csecond}.027 & -2.437 & \cellcolor{csecond}.030 \\
\textsc{Sealect} (kernel) & \cellcolor{csecond}-2.332 & \cellcolor{csecond}.027 & \cellcolor{csecond}-2.414 & .031 \\
\bottomrule
\end{tabular}
}
\caption{Plausibility (Joint-Pose log-likelihood and violation rate). \textsc{Sealect} more than halves the violations of appearance retrieval and improves likelihood over the discrete bin; prior-only attains the best plausibility only by collapsing diversity (cf.\ Table~\ref{tab:dist_compact}). Full per-axis results in supplementary Table~\ref{tab:plausibility_combined}. $^\dagger$Discrete-bin prior.}
\label{tab:plaus_compact}
\end{table}

\subsection{Comparison with Learned Compatibility Baselines}
\label{sec:learned_baselines}

\begin{table}[t]
\centering
\small
\setlength{\tabcolsep}{4pt}
\renewcommand{\arraystretch}{1.1}
\resizebox{\columnwidth}{!}{
\begin{tabular}{l ccc|ccc}
\toprule
& \multicolumn{3}{c|}{$K=20$} & \multicolumn{3}{c}{$K=100$} \\
\textbf{Geometry term} & Recall $\uparrow$ & NDCG $\uparrow$ & PS $\uparrow$ & Recall $\uparrow$ & NDCG $\uparrow$ & PS $\uparrow$ \\
\midrule
$S_{\text{neural}}$ only & \cellcolor{cbest}\textbf{0.634} & 0.830 & \cellcolor{csecond}2.278 & \cellcolor{cbest}\textbf{0.809} & 0.814 & \cellcolor{csecond}2.066 \\
\midrule
$S_{\text{geom}}$ only (geom.\ A)$^\dagger$ & \cellcolor{cworst}0.028 & \cellcolor{cbest}\textbf{0.940} & 1.812 & \cellcolor{cworst}0.039 & \cellcolor{cbest}\textbf{0.935} & 1.825 \\
GALA-style only~\cite{gala} (geom.\ B) & 0.107 & \cellcolor{cworst}0.813 & \cellcolor{cworst}1.672 & 0.194 & 0.812 & \cellcolor{cworst}1.695 \\
\midrule
$S_{\text{neural}}$ $+$ GALA-style (learned) & 0.588 & 0.816 & 2.227 & 0.775 & \cellcolor{cworst}0.805 & 2.047 \\
$S_{\text{neural}}$ $+$ $S_{\text{geom}}$ (\textsc{Sealect}) & \cellcolor{csecond}0.629 & \cellcolor{csecond}0.846 & \cellcolor{cbest}\textbf{2.408} & \cellcolor{csecond}0.800 & \cellcolor{csecond}0.830 & \cellcolor{cbest}\textbf{2.201} \\
\bottomrule
\end{tabular}
}
\caption{Controlled geometry-term swap. The trained neural score is held fixed and only the geometry term is swapped: an explicit generative prior (\textsc{Sealect}) versus a learned GALA-style discriminative compatibility scorer. Both are fused with the neural score under identical per-query $z$-normalization and no observation term ($\lambda_o{=}0$), so the \textsc{Sealect} row is the $(1,1,0)$ setting of supplementary Table~\ref{tab:lambda}. $^\dagger$Discrete-bin prior.}
\label{tab:gala_swap}
\end{table}

The scene-conditioned terms in \textsc{Sealect} are explicit generative priors rather than a learned discriminative compatibility model. We therefore compare against two learned foreground-object-search (FOS) baselines: an adapted DiscoFOS~\cite{discofos} model evaluated as a full system, and a GALA-style learned geometry term evaluated in a controlled swap. ``GALA-style'' is a discriminative scorer modeled on GALA~\cite{gala}, whose image encoders cannot take our pose-level inputs: an MLP over scene statistics and candidate pose, placement, and scale, trained on real versus mismatched pairs (also the standalone baseline in every table). Unlike the kernel prior, its scene input omits the target region's position and size.

\noindent\textbf{DiscoFOS.}
We fine-tune the official DiscoFOS student on the same scene--pose library, positives, and random-negative structure as our neural retriever, and give it the ground-truth target box (its best case); it ranks the full 68{,}338-candidate pool under the identical validity filter and metric as Table~\ref{tab:retrieval_combined}. Fine-tuning lifts it well above zero-shot (Recall@20 $0.071\!\rightarrow\!0.333$), but it still trails \textsc{Sealect}, the neural retriever, and the CLIP/DINO baselines on recall: seeing only the box-masked background, it cannot exploit their crop-to-crop appearance shortcut. Its lower NDCG and PS reflect weaker pose fidelity, and in supplementary Tables~\ref{tab:dist_exp2} and~\ref{tab:plausibility_combined} its yaw and joint-pose JS are close to \textsc{Sealect}'s, but it is markedly worse on vertical placement (Y-bin JS $0.184$ vs.\ $0.051$), scale ($0.085$ vs.\ $0.052$), and joint-pose violations ($0.075$ vs.\ $0.027$).

\noindent\textbf{GALA-style geometry swap.}
To isolate the geometry signal, we hold the trained neural score fixed and swap only the geometry term: the explicit generative prior (\textsc{Sealect}) versus the GALA-style scorer (Table~\ref{tab:gala_swap}). The explicit prior wins on all three metrics at both pool sizes (Recall $+0.041$/$+0.025$, NDCG $+0.030$/$+0.025$, PS $+0.181$/$+0.154$ at $K{=}20$/$100$), and the fused GALA-style variant has a higher joint-pose violation rate than full \textsc{Sealect} ($0.068$ vs.\ $0.027$; supplementary Table~\ref{tab:plausibility_combined}). Neither the full learned FOS system nor the learned geometry term matches the training-free explicit prior.

\subsection{Downstream Detector Precision}
\label{sec:downstream}

The evaluations above are intrinsic to the composites; we finally ask how insertion geometry affects a \emph{downstream} detector. We train Faster R-CNN~\cite{ren2015faster} detectors on a maritime base set (SeaShips\,+\,SMD) and paste vessels into each image under one of two geometries, with crops and paste centers shared across conditions through a common random seed: \emph{scene-consistent} (scale tied to vertical position, upright), the configuration \textsc{Sealect} favors, or \emph{implausible} (scale decoupled from position, rotated by up to $\pm35^\circ$ and randomly flipped), which appearance-only selection admits. We evaluate on the held-out LaRS distribution ($3$ seeds, SAM\,2-masked~\cite{ravi2024sam2} cutouts; supplementary Section~\ref{sec:downstream_supp}). In this setup copy-paste augmentation of either kind does \emph{not} help: relative to no augmentation, $\text{AP}_{50}$ drops ($0.186$ vs.\ $0.159$/$0.165$) and false detections per vessel-free image rise ($0.20$ vs.\ $0.54$ plausible, $0.67$ implausible). Implausible insertions produce more false positives than scene-consistent ones in all three paired seeds (${\sim}25\%$ more on average), while AP differs little, in line with the finding that context-agnostic pasting already yields AP gains~\cite{ghiasi2021simplee}. Implausible pastes are also larger and more often skipped when they do not fit, so the comparison is not fully controlled, and it does not show that \textsc{Sealect}-selected augmentation improves detection.

\subsection{Human Plausibility Study}
\label{sec:human}

To assess plausibility \emph{outside} our scene-conditioned prior (the concern that likelihood and violation metrics are scored under scene-conditioned statistics of the kind \textsc{Sealect} optimizes), we run a blind paired-comparison study over the end-to-end composites of six methods: each trial shows two composites of the \emph{same} background differing only in the inserted, outlined vessel, and raters pick the insertion that looks more realistic for the scene or answer ``can't tell''  (full protocol in supplementary Section~\ref{sec:human_protocol}). Over $1{,}110$ background-matched comparisons across all $15$ method pairs from six naive raters (Fleiss $\kappa\,{=}\,0.31$), a Bradley--Terry~\cite{bradley1952rank} fit ranks \textsc{Sealect} \emph{first} of six (Table~\ref{tab:human_arena}), unchanged when the blind author, advisor, and a partial rater are added (nine raters). The study uses the discrete-bin \textsc{Sealect} (Table~\ref{tab:retrieval_combined}). With intervals from resampling individual judgments, \textsc{Sealect} separates from random and GALA-style selection and is statistically tied with CLIP, DiscoFOS, and DINO; GALA-style even ranks \emph{below} random. The composites keep BootPlace's occasional off-water placements, as a deployed pipeline would. Raters thus judge \textsc{Sealect}'s composites at least as plausible as the best appearance retrieval, not more; its measured advantage is in the geometric statistics above.

\begin{table}[t]
\centering
\small
\setlength{\tabcolsep}{3.5pt}
\begin{tabular}{l c c c}
\toprule
Method & BT $\uparrow$ & $95\%$ CI & \textsc{Sealect} pref. \\
\midrule
\textsc{Sealect} & \cellcolor{cbest}\textbf{0.213} & $[0.188, 0.237]$ & -- \\
CLIP     & \cellcolor{csecond}$0.211$ & $[0.178, 0.251]$ & $57\%$ \\
DiscoFOS & $0.184$ & $[0.153, 0.223]$ & $52\%$ \\
DINO     & $0.168$ & $[0.138, 0.200]$ & $60\%$ \\
Random   & $0.125$ & $[0.100, 0.151]$ & $61\%$ \\
GALA-style & \cellcolor{cworst}$0.100$ & $[0.079, 0.121]$ & $64\%$ \\
\bottomrule
\end{tabular}
\caption{Blind human plausibility study ($1{,}110$ comparisons from six naive raters, Fleiss $\kappa{=}0.31$). Bradley--Terry scores (higher is better; ties count half) with $95\%$ CIs from resampling judgments, and \textsc{Sealect}'s head-to-head preference rate excluding ties. \textsc{Sealect} ranks first; its interval separates from random and GALA-style (significant) and overlaps CLIP/DiscoFOS/DINO (tied). The ranking is unchanged on all nine raters ($\kappa{=}0.31$).}
\label{tab:human_arena}
\end{table}

\section{Conclusion, Limitations and Future Work}

This paper has presented \textsc{Sealect}, a scene-conditioned retrieval framework that ranks vessel candidates by appearance, pose, and scene geometry jointly. Its recall comes from the trained retriever, while its training-free prior adds ranking quality and plausibility that learned alternatives do not match (Section~\ref{sec:learned_baselines}); online scoring takes ${\approx}12$\,ms per query on CPU.
\textbf{Limitations.} Likelihood and violation metrics use scene-conditioned statistics of the kind \textsc{Sealect} optimizes, PS shares the prior's pose estimator, and the prior-free human study shows parity with appearance retrieval, not superiority; augmentation did not beat no augmentation in our detector study. Leakage is controlled with embedding similarity rather than sequence labels, apparent pose is transferred without camera intrinsics, the contrastive objective can penalize valid alternatives, the observation term is uniform for $26\%$ of queries, and pose-estimator error caps the priors' fidelity; an implementation error in the symmetric yaw loss also concentrates half of the estimated yaw bins in one class, which coarsens every yaw-dependent metric (supplementary Section~\ref{sec:extra}).

\section*{Acknowledgments}
The authors thank the Johns Hopkins University Applied Physics Laboratory for supporting the first author's graduate studies through its part-time study benefits and for providing the in-house maritime data used in this work.

{
    \small
    \bibliographystyle{ieeenat_fullname}
    \bibliography{main}
}
\clearpage
\maketitlesupplementary

\section{Algorithmic Procedure}
Algorithm~\ref{alg:sealect} summarizes offline scene--pose library construction and online kernel-conditioned retrieval.
\begin{algorithm}[h]
\caption{\textsc{Sealect}: Scene-Conditioned Retrieval}
\label{alg:sealect}
\small
\begin{algorithmic}[1]
\Statex \textbf{Input:} Annotated corpus $\mathcal{I}=\{(I_i,\mathcal{B}_i)\}$, query background $x_{\mathrm{bg}}$, observed boxes $\mathcal{B}_{\mathrm{obs}}$, pose estimator $f_\phi$, pose discretizer $b(\cdot)$, segmentation model $\mathrm{Seg}$, scene-descriptor extractor $g$, scene encoder $f_{\text{scene}}$, object encoder $f_{\text{obj}}$, kernel bandwidth $\sigma$, weights $(\lambda_n,\lambda_g,\lambda_o)$
\Statex \textbf{Output:} Top-$K$ scene-compatible vessel candidates $\mathcal{O}^{\star}$

\Statex \textbf{Offline library construction}
\State Initialize candidate library $\mathcal{O}\gets\emptyset$
\For{$(I_i,\mathcal{B}_i) \in \mathcal{I}$}
    \State $\mathbf{c}_i \gets g(\mathrm{Seg}(I_i),\mathcal{B}_i)$ \Comment{scene statistics with vessel regions suppressed}
    \For{$b_{ij} \in \mathcal{B}_i$}
        \State $o_{ij} \gets I_i[b_{ij}]$, \quad $R_{ij} \gets f_\phi(o_{ij})$
        \State $u_{ij} \gets (b_{p,ij}, b_{y,ij}, b_{r,ij})$
        \State $y_{b,ij} \gets y(b_{ij})$, \quad $a_{ij} \gets a(b_{ij})$
        \State $\pi_{ij} \gets (u_{ij},y_{b,ij},a_{ij})$, \quad $e_{ij} \gets f_{\text{obj}}(o_{ij})$
        \State $\mathcal{O} \gets \mathcal{O} \cup \{(o_{ij},e_{ij},u_{ij},\pi_{ij},\mathbf{c}_i)\}$
    \EndFor
\EndFor
\State Retain per-entry geometric states $\{\pi_{ij}\}$ and scene descriptors $\{\mathbf{c}_i\}$ for kernel conditioning

\Statex \textbf{Scene-conditioned retrieval}
\State $\mathbf{c}_q \gets g(\mathrm{Seg}(x_{\mathrm{bg}}),\mathcal{B}_{\mathrm{obs}})$, \quad $q \gets f_{\text{scene}}(\mathbf{c}_q)$
\State $\mathcal{U}_{\mathrm{obs}} \gets \{b(f_\phi(x_{\mathrm{bg}}[\tilde{b}])) \mid \tilde{b} \in \mathcal{B}_{\mathrm{obs}}\}$
\State Estimate $C_{\mathrm{obs}}(u\mid\mathcal{U}_{\mathrm{obs}})$
\State $w_i \gets \operatorname{softmax}_i\!\big(-\lVert\mathbf{c}_q-\mathbf{c}_i\rVert^2/2\sigma^2\big)$ \Comment{kernel weights over library}
\For{$(o_j,e_j,u_j,\pi_j) \in \mathcal{O}$}
    \State $S_n(j) \gets q^\top e_j$
    \State $S_g(j) \gets \sum_f \beta_f \log \hat p_f(\pi_j \mid \mathbf{c}_q)$ \Comment{weighted factor log-likelihoods}
    \State $S_o(j) \gets \log C_{\mathrm{obs}}(u_j\mid\mathcal{U}_{\mathrm{obs}})$
\EndFor
\State $z$-normalize $S_n(\cdot),S_g(\cdot),S_o(\cdot)$ across candidates \Comment{per query}
\State $S(j) \gets \lambda_n S_n(j)+\lambda_g S_g(j)+\lambda_o S_o(j)$ for all $j$
\State $\mathcal{O}^{\star} \gets \operatorname{TopK}_{(o_j,\cdot)\in\mathcal{O}} S(j)$
\State \Return $\mathcal{O}^{\star}$
\end{algorithmic}
\end{algorithm}

\section{Implementation Details}
\label{sec:impl}
\subsection{Neural Retriever} The scene encoder $f_{\text{scene}}$ and candidate encoder $f_{\text{obj}}$ are two-tower MLPs with the same architecture, $\mathrm{Linear}(\cdot,512)\rightarrow\mathrm{ReLU}\rightarrow\mathrm{LayerNorm}\rightarrow\mathrm{Dropout}(0.05)\rightarrow\mathrm{Linear}(512,256)\rightarrow\mathrm{ReLU}\rightarrow\mathrm{LayerNorm}\rightarrow\mathrm{Linear}(256,128)$, yielding $\ell_2$-normalized $128$-d embeddings. The scene tower ingests background scene statistics (water/sky/land/other and vessel-masked area fractions, normalized horizon location and its confidence, per-class row mean/std, and $16$-bin vertical class profiles) together with the target region's local placement geometry (normalized center, vertical-placement bin, $\log$-area, aspect ratio) and one-hot viewpoint/layout descriptors. The candidate tower ingests the crop's $384$-d DINOv2 ViT-S/14~\cite{oquab2024dinov2} embedding ($\ell_2$-normalized) and the pose estimator's $256$-d trunk embedding, concatenated with the candidate's pose bins, vertical-placement bin, $\log$-area, aspect ratio, normalized center, its source-scene background statistics, and one-hot viewpoint, layout, pose-family, and yaw-family descriptors. The neural term is therefore a \emph{learned scene--candidate compatibility} over appearance, scene context, and placement geometry, while the explicit prior $S_{\text{geom}}$ contributes complementary, training-free geometric structure on top. The towers are trained jointly with a symmetric in-batch InfoNCE~\cite{oord2018cpc,radford2021learning} objective (temperature $\tau{=}0.07$) using AdamW (learning rate $10^{-3}$, weight decay $10^{-4}$), batch size $256$, for $30$ epochs (seed $42$); DINO embeddings are precomputed. At inference $S_{\text{neural}}(j)=q^\top e_j$.

\subsection{Scene--Pose Library} The scene descriptor $\mathbf{c}$ is derived from a grounded segmentation map $\mathrm{Seg}(I)$, in which Grounding DINO~\cite{liu2024groundingdino} boxes for water, sky, and land are converted to masks by SAM\,2~\cite{ravi2024sam2} (regions left unassigned by the grounding are completed with a color--position heuristic), as the fractions of pixels labeled water, sky, and land together with a $B$-band vertical profile; annotated vessel boxes are masked before averaging so $\mathbf{c}$ captures background context only. Per vessel the pose bins $u=(b_p,b_y,b_r)$ are the argmax outputs of the estimator's bin classifiers; we also discretize the normalized vertical center $y_c$ into a placement bin $y_b=\lfloor y_c N_y\rfloor$, and record relative scale $a=|b|/|I|$. The context key $\kappa(\mathbf{c})$ combines the viewpoint class, the layout class, five-level quantizations of the water, sky, and land fractions, and a five-level horizon bucket; $313$ keys occur in the training split. It is used by the discrete-bin variant (the ablation of Table~\ref{tab:retrieval_combined}, the rows marked $\dagger$, and the rendered and human evaluations) and to index the held-out evaluation reference.

\subsection{Priors and Fusion} The scene-conditioned prior (Eq.~\ref{eq:kernel_prior}) is fully empirical (kernel-weighted histograms for the pose and placement factors and a kernel-weighted Gaussian over log-scale, with bandwidth $\sigma{=}1.0$ over standardized scene features), factorized into pitch/yaw/roll pose marginals, vertical placement, and log-scale, and combined with fixed factor weights ($2{:}1{:}1$ for pitch:yaw:roll, $0.7$ for scale, $0.5$ for placement) that emphasize the reliably estimated pitch axis. Retrieval fuses the three terms under per-query $z$-normalization with $(\lambda_n,\lambda_g,\lambda_o){=}(1,1,0.10)$ over a pool of $68{,}338$ candidates. The discrete-bin variant differs in more than the weighting: its histograms are weighted by pose confidence, its yaw weight adapts to the entropy of the bin's yaw histogram, its scale deviation is floored at $0.15$, and the $5.5\%$ of test queries whose key is absent from the training split fall back to a five-class scene-type prior.

\subsection{Datasets} The retrieval corpus spans seven maritime benchmarks: ABOShips~\cite{iancu2021aboshipsinshoreoffshore}, KOLOMVERSE~\cite{10689331}, LaRS~\cite{zust2023larsdiversepanopticmaritime}, SMD~\cite{prasad2017video}, SeaDronesSee~\cite{varga2022seadronessee} (SDS), SeaShips~\cite{8438999}, and an access-restricted in-house UAV set (IH-UAV) provided by the Johns Hopkins University Applied Physics Laboratory; the first six are public. The pose estimator is additionally validated on MARVEL~\cite{MARVEL}, yielding the eight datasets used for pose evaluation.

\subsection{Splits and Leakage Control} We partition the retrieval corpus into a training candidate pool ($68{,}338$ vessels) and a test set ($4{,}245$ query crops, about $5.6\%$ of all $75{,}878$ crops) with strict image-level exclusivity, so no image contributes to both. Because the constituent benchmarks lack consistent temporal/sequence labels, we control near-duplicates with DINOv2 ViT-S/14 embeddings instead: crops are grouped into near-duplicate clusters (cosine $\geq0.95$) that stay on one side of the split, and training images whose mean embedding has cosine $\geq0.94$ to a test image are removed ($630$ images, $3{,}295$ crops). Test images are sampled per scene-context bin ($5\%$ per bin, clamped to $15$--$120$ crops, seed $42$) and then expanded to whole near-duplicate clusters. During retrieval, candidates from the query's own image or from the same filename-derived sequence group are excluded; because KOLOMVERSE filenames carry no sequence information, this removes all KOLOMVERSE candidates for KOLOMVERSE queries.

\section{Evaluation Metrics}
\label{sec:eval_metrics}

We evaluate retrieval quality, distributional alignment, and scene-conditioned plausibility using both ranking-based and probabilistic metrics. Metrics are computed over retrieved candidate sets under constrained ($K{=}20$) and expanded ($K{=}100$) retrieval regimes.

\subsection{Retrieval Quality}

To evaluate instance retrieval performance, we report Recall@K and normalized discounted cumulative gain (NDCG). Recall measures whether at least one compatible candidate is retrieved within the top-$K$ results:
\begin{equation}
\mathrm{Recall@}K
=
\frac{1}{N}
\sum_{i=1}^{N}
\mathbf{1}
\big[
\exists j \le K :
o_j \in \mathcal{R}_i
\big],
\end{equation}
where $\mathcal{R}_i$ denotes the set of valid candidates for query scene $i$. Ranking quality is evaluated using NDCG:
\begin{equation}
\mathrm{NDCG@}K
=
\frac{1}{Z}
\sum_{j=1}^{K}
\frac{2^{r_j}-1}{\log_2(j+1)},
\end{equation}
where $Z$ is the DCG of the same top-$K$ list re-sorted by relevance, so NDCG measures how well a method orders the candidates it retrieves and can be high for a method that retrieves few fully relevant candidates (\eg the $S_{\text{geom}}$-only ablation). The graded relevance $r_j\in\{0,\dots,4\}$ counts the geometric attributes on which candidate $j$ matches the query scene's held-out vessel: the same pitch bin, yaw within one $20^\circ$ bin (circular), the same vertical-placement bin, and $|\log a_j-\log a_q|\le0.30$; the valid set $\mathcal{R}_i$ holds the candidates with $r_j{=}4$. To quantify geometric consistency, the plausibility score (PS) is the mean relevance of the retrieved list,
\begin{equation}
\mathrm{PS}
=
\frac{1}{K}
\sum_{j=1}^{K}
r_j \;\in[0,4],
\end{equation}
the average number of attributes per retrieved candidate that agree with the held-out vessel. PS is computed against the held-out vessel's geometry rather than \textsc{Sealect}'s prior, although both use the same pose estimator; the plausibility metrics below instead score against a held-out empirical reference $p$.

\subsection{Distributional Alignment}

To evaluate whether retrieved candidates preserve the empirical maritime pose and placement distributions, we compare retrieved distributions against ground-truth scene-conditioned distributions using Jensen--Shannon (JS) divergence, computed per context bin with base-2 logarithms and averaged over bins with at least $20$ ground-truth and $20$ retrieved samples:
\begin{equation}
\mathrm{JS}(P \,\|\, Q)
=
\frac{1}{2}
\mathrm{KL}(P \,\|\, M)
+
\frac{1}{2}
\mathrm{KL}(Q \,\|\, M),
\end{equation}
where
\(
M = \frac{1}{2}(P+Q)
\)
and $\mathrm{KL}(\cdot)$ denotes Kullback--Leibler divergence.

We additionally report entropy of the retrieved distribution:
\begin{equation}
H_{\mathrm{ret}}
=
-
\sum_u
p_{\mathrm{ret}}(u)
\log p_{\mathrm{ret}}(u),
\end{equation}
which measures retrieval spread across pose or placement configurations. Lower entropy indicates concentrated retrieval behavior, while higher entropy reflects broader configuration coverage. To quantify how far retrieval spread departs from the empirical spread, we report the entropy gap, averaged over the same context bins:
\begin{equation}
\mathrm{Gap}
=
\left|
H(p_{\mathrm{ret}})
-
H(p_{\mathrm{gt}})
\right|
\quad\text{(bits)},
\end{equation}
where $p_{\mathrm{ret}}$ and $p_{\mathrm{gt}}$ denote retrieved and empirical scene-conditioned distributions respectively.

\subsection{Pose and Placement Plausibility}

To measure physical plausibility \emph{externally}, we evaluate the likelihood of retrieved configurations under a held-out empirical reference $p(\cdot\mid\gamma)$ estimated from ground-truth test scenes (histograms over the discrete context key $\gamma$ with add-$\frac{1}{2}$ smoothing). Each axis $x\in\{$yaw, joint pose $u$, y-bin, scale bin$\}$ is scored separately; the ``Joint'' columns use the pose bins $u$ only. Unlike PS, this is a proper discrete log-probability and is therefore bounded above by zero:
\begin{equation}
\mathcal{L}_{\mathrm{pose}}
=
\frac{1}{K}
\sum_{j=1}^{K}
\log p(x_j \mid \gamma).
\end{equation}

We further compute violation rates corresponding to low-probability configurations:
\begin{equation}
\mathrm{Viol.}
=
\frac{1}{K}
\sum_{j=1}^{K}
\mathbf{1}
\big[
p(x_j \mid \gamma)
<
\tau
\big],
\end{equation}
with threshold $\tau{=}0.01$. Violation rates capture physically inconsistent configurations such as implausible vessel orientation, incorrect horizon-relative scale, or unrealistic vertical placement. Lower violation rates therefore indicate stronger alignment between retrieved candidates and scene-conditioned maritime geometry.

\subsection{Rendered Compatibility Evaluation}

For the rendered pipeline (Section~\ref{sec:rendered}), each inserted candidate receives
\begin{equation}
\begin{split}
S_{\mathrm{compat}} ={}& \textstyle\sum_{m} w_m\,\mathrm{clip}\big(1+\log p(x_m\mid\gamma)/\log|V_m|,\,0,\,1\big)\\
&- 0.05\,n_{\mathrm{viol}} - 0.5\,\mathbf{1}_{\mathrm{overlap}} - 0.75\,\mathbf{1}_{\mathrm{no\,render}},
\end{split}
\end{equation}
where $m$ ranges over joint pose ($w_m{=}0.35$), y-bin ($0.25$), scale bin ($0.25$), and yaw ($0.15$), $p(\cdot\mid\gamma)$ is the held-out reference above, $|V_m|$ is the number of bins of axis $m$, $n_{\mathrm{viol}}$ counts axes with $p<0.01$, and the indicators mark insertions rejected for overlap or missing a rendered composite. The geometric tokens and $\gamma$ come from the inserted candidate's source metadata. We also report mean rank and top-1 frequency among the methods rendered for each scene.

\section{Full Distributional and Plausibility Results}
The complete per-axis distributional-alignment and plausibility results summarized in the main paper (Section~\ref{sec:experiments}) are reported in Tables~\ref{tab:dist_exp2} and~\ref{tab:plausibility_combined}.

\begin{table*}[t]
\centering
\small
\setlength{\tabcolsep}{4pt}
\renewcommand{\arraystretch}{1.08}
\resizebox{0.95\linewidth}{!}{
\begin{tabular}{l c ccc ccc ccc ccc}
\toprule &
& \multicolumn{3}{c}{Yaw}
& \multicolumn{3}{c}{Joint Pose}
& \multicolumn{3}{c}{Y-bin}
& \multicolumn{3}{c}{Scale} \\
\cmidrule(lr){3-5}\cmidrule(lr){6-8}\cmidrule(lr){9-11}\cmidrule(lr){12-14}
Method & $K$ & JS$\downarrow$ & Gap$\downarrow$ & $H_{\text{ret}}$
& JS$\downarrow$ & Gap$\downarrow$ & $H_{\text{ret}}$
& JS$\downarrow$ & Gap$\downarrow$ & $H_{\text{ret}}$
& JS$\downarrow$ & Gap$\downarrow$ & $H_{\text{ret}}$ \\
\midrule

Random & 20  & 0.0812 & 0.4172 & 1.9089 & 0.2410 & 0.9688 & 3.7151 & 0.2258 & 0.5917 & 2.3623 & 0.1612 & 0.5103 & 2.5549 \\
Random & 100 & 0.0798 & 0.4174 & 1.9136 & 0.2377 & 1.0157 & 3.7751 & 0.2239 & 0.5938 & 2.3654 & 0.1615 & 0.5090 & 2.5582 \\
\midrule

CLIP & 20  & \cellcolor{csecond}0.0521 & \cellcolor{csecond}0.2570 & 1.7321 & \cellcolor{csecond}0.1703 & 0.5932 & 3.3490 & 0.1730 & 0.5420 & 2.3622 & 0.0730 & 0.2991 & 2.3239 \\
CLIP & 100 & \cellcolor{csecond}0.0556 & 0.2833 & 1.7797 & 0.1797 & 0.6989 & 3.4654 & 0.1869 & 0.5631 & 2.3875 & 0.0880 & 0.3364 & 2.3475 \\
\midrule

DINO & 20  & \cellcolor{cbest}\textbf{0.0508} & \cellcolor{cbest}\textbf{0.2312} & 1.7065 & \cellcolor{cbest}\textbf{0.1591} & \cellcolor{csecond}0.5307 & 3.2935 & 0.1707 & 0.5001 & 2.3201 & 0.0766 & 0.3183 & 2.3303 \\
DINO & 100 & \cellcolor{cbest}\textbf{0.0528} & \cellcolor{cbest}\textbf{0.2662} & 1.7495 & \cellcolor{cbest}\textbf{0.1660} & 0.6392 & 3.4079 & 0.1862 & 0.5130 & 2.3369 & 0.0894 & 0.3754 & 2.3803 \\
\midrule

DINO + $S_{\text{geom}}^\dagger$
& 20  & 0.0636 & 0.3680 & 1.3115 & 0.1974 & 0.6975 & 2.2538 & 0.1511 & 0.3537 & 1.9825 & 0.0811 & 0.2610 & 2.2063 \\
DINO + $S_{\text{geom}}^\dagger$
& 100 & 0.0620 & 0.3478 & 1.3411 & 0.1965 & 0.6455 & 2.3467 & 0.1629 & 0.3765 & 2.0054 & 0.0902 & 0.2896 & 2.2518 \\
\midrule

$S_{\text{geom}}^\dagger$ $(\lambda_g{=}1)$ 
& 20  & \cellcolor{cworst}0.3036 & \cellcolor{cworst}1.4664 & 0.0856 
& \cellcolor{cworst}0.5485 & \cellcolor{cworst}2.6589 & 0.1166 
& \cellcolor{cworst}0.5233 & \cellcolor{cworst}1.6952 & 0.1392 
& \cellcolor{cworst}0.5914 & \cellcolor{cworst}1.9040 & 0.1562 \\
$S_{\text{geom}}^\dagger$ 
& 100 & \cellcolor{cworst}0.2976 & \cellcolor{cworst}1.4514 & 0.0912 
& \cellcolor{cworst}0.5393 & \cellcolor{cworst}2.6250 & 0.1505 
& \cellcolor{cworst}0.4665 & \cellcolor{cworst}1.6125 & 0.2219 
& \cellcolor{cworst}0.5405 & \cellcolor{cworst}1.7661 & 0.3051 \\

\midrule

$S_{\text{neural}}$ $(\lambda_n{=}1)$
& 20  & 0.0609 & 0.3167 & 1.7110 & 0.1845 & 0.5476 & 3.2756 & \cellcolor{csecond}0.0620 & 0.2601 & 2.0766 & \cellcolor{cbest}\textbf{0.0504} & 0.2441 & 2.2827 \\
$S_{\text{neural}}$
& 100 & 0.0594 & 0.3177 & 1.7638 & 0.1865 & 0.6830 & 3.4303 & \cellcolor{csecond}0.0692 & 0.3347 & 2.1680 & \cellcolor{csecond}0.0577 & 0.2992 & 2.3424 \\

\midrule

$S_{\text{obs}}$ $(\lambda_o{=}1)$
& 20  & 0.1146 & 0.4745 & 1.1436 & 0.2400 & 0.9598 & 1.8465 & 0.4035 & 0.8014 & 1.0715 & 0.2150 & 0.3357 & 2.0398 \\
$S_{\text{obs}}$
& 100 & 0.1138 & 0.4814 & 1.1236 & 0.2401 & 1.0596 & 1.7324 & 0.3807 & 0.7483 & 1.1519 & 0.1899 & 0.3730 & 2.1856 \\

\midrule

DiscoFOS~\cite{discofos} (ft)
& 20  & 0.0713 & 0.2745 & 1.6898 & 0.1956 & 0.5339 & 3.2125 & 0.1844 & 0.4386 & 2.2267 & 0.0847 & 0.3015 & 2.2795 \\
DiscoFOS (ft)
& 100 & 0.0643 & \cellcolor{csecond}0.2759 & 1.7336 & 0.1859 & 0.6163 & 3.3659 & 0.1855 & 0.4627 & 2.2668 & 0.0864 & 0.3302 & 2.3283 \\

GALA-style only~\cite{gala}
& 20  & 0.2493 & 0.5072 & 1.1587 & 0.4439 & 0.9226 & 1.9930 & 0.3171 & 0.7669 & 1.2299 & 0.2893 & 0.3585 & 1.9247 \\
GALA-style only
& 100 & 0.1748 & 0.3793 & 1.4106 & 0.3537 & 0.6671 & 2.3855 & 0.3032 & 0.7232 & 1.2825 & 0.2472 & 0.4027 & 2.0342 \\

$S_{\text{neural}}$ + GALA-style
& 20  & 0.0688 & 0.3132 & 1.6941 & 0.2055 & 0.5350 & 3.1592 & 0.0864 & 0.3075 & 1.9200 & 0.0602 & 0.2562 & 2.2518 \\
$S_{\text{neural}}$ + GALA-style
& 100 & 0.0666 & 0.3195 & 1.7429 & 0.2002 & 0.6333 & 3.3204 & 0.0965 & 0.3423 & 1.9691 & 0.0731 & 0.3154 & 2.3067 \\

\midrule

\textsc{Sealect} (discrete)
& 20  & 0.0629 & 0.3407 & 1.3816 & 0.1983 & 0.5595 & 2.4585 & 0.0668 & \cellcolor{csecond}0.1908 & 1.9284 & 0.0526 & \cellcolor{cbest}\textbf{0.2161} & 2.2416 \\
\textsc{Sealect} (discrete)
& 100 & 0.0591 & 0.3163 & 1.4508 & 0.1913 & \cellcolor{csecond}0.4990 & 2.6425 & 0.0728 & \cellcolor{csecond}0.2262 & 1.9878 & 0.0603 & \cellcolor{csecond}0.2626 & 2.2833 \\
\textsc{Sealect} (kernel)
& 20  & 0.0617 & 0.3857 & 1.2499 & 0.1846 & \cellcolor{cbest}\textbf{0.5251} & 2.3986 & \cellcolor{cbest}\textbf{0.0515} & \cellcolor{cbest}\textbf{0.1833} & 1.9647 & \cellcolor{csecond}0.0520 & \cellcolor{csecond}0.2188 & 2.2366 \\
\textsc{Sealect} (kernel)
& 100 & 0.0558 & 0.3141 & 1.3315 & \cellcolor{csecond}0.1776 & \cellcolor{cbest}\textbf{0.4153} & 2.5944 & \cellcolor{cbest}\textbf{0.0581} & \cellcolor{cbest}\textbf{0.2158} & 2.0163 & \cellcolor{cbest}\textbf{0.0565} & \cellcolor{cbest}\textbf{0.2574} & 2.2726 \\

\bottomrule
\end{tabular}
}
\caption{
Distributional alignment of retrieved maritime vessel configurations across pose, placement, and scale. \textsc{Sealect} achieves the best tradeoff between entropy alignment and distributional fidelity, correcting neural over-dispersion while avoiding prior collapse. Shading (best green, second yellow, worst red) is per column within each $K$; $H_{\text{ret}}$ is unshaded. $^\dagger$Discrete-bin prior.
}
\label{tab:dist_exp2}
\end{table*}

\begin{table*}[t]
\centering
\small
\setlength{\tabcolsep}{8pt}
\renewcommand{\arraystretch}{0.95}
\resizebox{0.75\textwidth}{!}{
\begin{tabular}{l c cccc cccc}
\toprule
& & \multicolumn{4}{c}{Log-Likelihood $\uparrow$} & \multicolumn{4}{c}{Violation Rate $\downarrow$} \\
\cmidrule(lr){3-6} \cmidrule(lr){7-10}
Method & $K$ & Yaw & Joint & Y-bin & Scale & Yaw & Joint & Y-bin & Scale \\
\midrule

Random & 20  & -1.477 & \cellcolor{cworst}-3.053 & \cellcolor{cworst}-2.188 & -2.197 & \cellcolor{cworst}.024 & \cellcolor{cworst}.120 & \cellcolor{cworst}.060 & .028 \\
Random & 100 & \cellcolor{cworst}-1.475 & \cellcolor{cworst}-3.054 & \cellcolor{cworst}-2.183 & \cellcolor{cworst}-2.197 & \cellcolor{cworst}.024 & \cellcolor{cworst}.119 & \cellcolor{cworst}.060 & .027 \\

CLIP & 20  & -1.290 & -2.747 & -1.983 & -1.739 & .012 & .073 & .033 & .010 \\
CLIP & 100 & -1.330 & -2.813 & -2.043 & -1.806 & .014 & .081 & .037 & .013 \\

DINO & 20  & -1.274 & -2.699 & -1.948 & -1.762 & .013 & .066 & .029 & .010 \\
DINO & 100 & -1.304 & -2.751 & -2.007 & -1.841 & .014 & .074 & .031 & .013 \\

DINO+$S_{\text{geom}}^\dagger$ & 20  & -1.048 & -2.248 & -1.648 & -1.711 & \cellcolor{csecond}.003 & \cellcolor{csecond}.021 & .014 & \cellcolor{cbest}\textbf{.005} \\
DINO+$S_{\text{geom}}^\dagger$ & 100 & \cellcolor{csecond}-1.066 & -2.289 & -1.704 & -1.775 & \cellcolor{csecond}.003 & \cellcolor{csecond}.023 & .016 & \cellcolor{csecond}.007 \\

\midrule

$S_{\text{geom}}^\dagger$ & 20  & \cellcolor{cbest}\textbf{-0.880} & \cellcolor{cbest}\textbf{-1.963} & -1.471 & -1.740 & \cellcolor{cbest}\textbf{.000} & \cellcolor{cbest}\textbf{.006} & .031 & .020 \\
$S_{\text{geom}}^\dagger$ & 100 & \cellcolor{cbest}\textbf{-0.876} & \cellcolor{cbest}\textbf{-1.968} & \cellcolor{cbest}\textbf{-1.396} & -1.711 & \cellcolor{cbest}\textbf{.000} & \cellcolor{cbest}\textbf{.006} & .026 & .013 \\

$S_{\text{neural}}$ & 20  & -1.306 & -2.742 & -1.491 & -1.666 & .012 & .066 & .007 & .007 \\
$S_{\text{neural}}$ & 100 & -1.336 & -2.787 & -1.563 & -1.716 & .014 & .069 & .010 & .009 \\

$S_{\text{obs}}$ & 20  & -1.084 & \cellcolor{csecond}-2.233 & -1.771 & -2.030 & .006 & .027 & \cellcolor{csecond}.004 & .012 \\
$S_{\text{obs}}$ & 100 & -1.075 & \cellcolor{csecond}-2.185 & -1.758 & -2.043 & .006 & .025 & \cellcolor{cbest}\textbf{.004} & .009 \\

\midrule

DiscoFOS~\cite{discofos} (ft) & 20  & -1.339 & -2.790 & -1.894 & -1.824 & .014 & .075 & .029 & .024 \\
DiscoFOS (ft) & 100 & -1.344 & -2.806 & -1.951 & -1.807 & .014 & .077 & .033 & .016 \\

GALA-style only~\cite{gala} & 20  & \cellcolor{cworst}-1.485 & -2.861 & -1.800 & \cellcolor{cworst}-2.261 & .015 & .083 & .025 & \cellcolor{cworst}.041 \\
GALA-style only & 100 & -1.443 & -2.817 & -1.785 & -2.161 & .012 & .081 & .024 & \cellcolor{cworst}.035 \\

$S_{\text{neural}}$ + GALA-style & 20  & -1.314 & -2.744 & -1.465 & -1.681 & .011 & .068 & \cellcolor{cbest}\textbf{.003} & \cellcolor{csecond}.006 \\
$S_{\text{neural}}$ + GALA-style & 100 & -1.329 & -2.771 & -1.531 & -1.738 & .012 & .068 & \cellcolor{csecond}.005 & .008 \\

\midrule

\textsc{Sealect} (discrete) & 20  & -1.096 & -2.354 & \cellcolor{csecond}-1.406 & \cellcolor{csecond}-1.647 & .005 & .027 & \cellcolor{cbest}\textbf{.003} & \cellcolor{csecond}.006 \\
\textsc{Sealect} (discrete) & 100 & -1.142 & -2.437 & -1.459 & \cellcolor{csecond}-1.690 & .006 & .030 & \cellcolor{cbest}\textbf{.004} & \cellcolor{csecond}.007 \\
\textsc{Sealect} (kernel) & 20  & \cellcolor{csecond}-1.040 & -2.332 & \cellcolor{cbest}\textbf{-1.395} & \cellcolor{cbest}\textbf{-1.631} & .004 & .027 & \cellcolor{cbest}\textbf{.003} & \cellcolor{csecond}.006 \\
\textsc{Sealect} (kernel) & 100 & -1.086 & -2.414 & \cellcolor{csecond}-1.446 & \cellcolor{cbest}\textbf{-1.665} & .005 & .031 & \cellcolor{cbest}\textbf{.004} & \cellcolor{cbest}\textbf{.006} \\

\bottomrule
\end{tabular}
}
\caption{
Plausibility evaluation across maritime vessel configurations. \textsc{Sealect} improves joint-pose likelihood and reduces violations relative to neural retrieval, consistently shifting samples toward high-probability regions while suppressing implausible configurations. Shading (best green, second yellow, worst red) is per column within each $K$. $^\dagger$Discrete-bin prior.
}
\label{tab:plausibility_combined}
\end{table*}

\section{Robustness and Determinism}
\subsection{Kernel Bandwidth}
We fix the kernel bandwidth to $\sigma{=}1.0$ (unit width on standardized scene features) \emph{a priori}, rather than tuning it on the test corpus. Table~\ref{tab:bandwidth} sweeps $\sigma\in\{0.5,1,2\}$: the continuous prior improves NDCG and PS over the discrete-bin ablation at \emph{every} bandwidth, and improves recall for $\sigma\le1$. NDCG and PS increase while recall decreases monotonically with $\sigma$, so $\sigma{=}1.0$ balances instance recall against ranking and plausibility. The kernel prior is a deterministic empirical estimate evaluated over a frozen encoder, so each configuration is exactly reproducible and the kernel-vs-discrete comparison carries no seed variance ($\sigma{=}1.0$ reproduces the main-paper \textsc{Sealect} rows of Tables~\ref{tab:retrieval_combined},~\ref{tab:dist_exp2}, and~\ref{tab:plausibility_combined} to all reported digits).

\begin{table}[t]
\centering
\small
\setlength{\tabcolsep}{4pt}
\resizebox{\columnwidth}{!}{
\begin{tabular}{l ccc|ccc}
\toprule
& \multicolumn{3}{c|}{$K=20$} & \multicolumn{3}{c}{$K=100$} \\
\textbf{Scene prior} & Recall $\uparrow$ & NDCG $\uparrow$ & PS $\uparrow$ & Recall $\uparrow$ & NDCG $\uparrow$ & PS $\uparrow$ \\
\midrule
Discrete bin (ablation) & 0.598 & \cellcolor{cworst}0.843 & \cellcolor{cworst}2.354 & \cellcolor{cworst}0.767 & \cellcolor{cworst}0.827 & \cellcolor{cworst}2.142 \\
\midrule
Kernel, $\sigma{=}0.5$ & \cellcolor{cbest}\textbf{0.634} & 0.846 & 2.407 & \cellcolor{cbest}\textbf{0.805} & 0.830 & 2.199 \\
Kernel, $\sigma{=}1.0$ (default) & \cellcolor{csecond}0.627 & \cellcolor{csecond}0.848 & \cellcolor{csecond}2.418 & \cellcolor{csecond}0.801 & \cellcolor{csecond}0.832 & \cellcolor{csecond}2.212 \\
Kernel, $\sigma{=}2.0$ & \cellcolor{cworst}0.596 & \cellcolor{cbest}\textbf{0.852} & \cellcolor{cbest}\textbf{2.420} & 0.774 & \cellcolor{cbest}\textbf{0.834} & \cellcolor{cbest}\textbf{2.214} \\
\bottomrule
\end{tabular}
}
\caption{Kernel bandwidth robustness. The continuous scene-conditioned prior beats the discrete-bin ablation on NDCG and PS at all bandwidths and on recall for $\sigma\le1$; the fixed operating point $\sigma{=}1.0$ is not finely tuned to the test set. Numbers are deterministic.}
\label{tab:bandwidth}
\end{table}

\subsection{Fusion-Weight Sensitivity} We fix the fusion weights to $(\lambda_n,\lambda_g,\lambda_o){=}(1,1,0.10)$ rather than tuning them per split. Table~\ref{tab:lambda} perturbs each weight around this point. At $K{=}20$ recall stays in $[0.600,0.637]$, NDCG in $[0.843,0.853]$, and PS in $[2.384,2.453]$, all comfortably above every appearance and learned-FOS baseline (Table~\ref{tab:retrieval_combined}). The trade-off is interpretable: up-weighting the geometric prior ($\lambda_g:0.5\!\rightarrow\!2$) exchanges recall for ranking and plausibility, while the observation term $\lambda_o$ contributes a small, consistent plausibility gain. \textsc{Sealect} is thus not finely tuned to the chosen weights.

\begin{table}[t]
\centering
\small
\setlength{\tabcolsep}{4pt}
\resizebox{\columnwidth}{!}{
\begin{tabular}{l ccc|ccc}
\toprule
& \multicolumn{3}{c|}{$K=20$} & \multicolumn{3}{c}{$K=100$} \\
$(\lambda_n,\lambda_g,\lambda_o)$ & Recall $\uparrow$ & NDCG $\uparrow$ & PS $\uparrow$ & Recall $\uparrow$ & NDCG $\uparrow$ & PS $\uparrow$ \\
\midrule
$(1,1,0.10)$ (default) & 0.627 & 0.848 & 2.418 & 0.801 & 0.832 & 2.212 \\
\midrule
$(1,0.5,0.10)$ & \cellcolor{cbest}\textbf{0.637} & \cellcolor{cworst}0.843 & \cellcolor{cworst}2.384 & \cellcolor{cbest}\textbf{0.813} & \cellcolor{cworst}0.827 & \cellcolor{cworst}2.170 \\
$(1,2.0,0.10)$ & \cellcolor{cworst}0.600 & \cellcolor{cbest}\textbf{0.853} & \cellcolor{cbest}\textbf{2.453} & \cellcolor{cworst}0.763 & \cellcolor{cbest}\textbf{0.838} & \cellcolor{cbest}\textbf{2.261} \\
$(1,1,0.0)$ & 0.629 & 0.846 & 2.408 & 0.800 & 0.830 & 2.201 \\
$(1,1,0.25)$ & 0.623 & \cellcolor{csecond}0.850 & 2.430 & 0.797 & 0.834 & 2.224 \\
$(0.75,1,0.10)$ & 0.617 & \cellcolor{csecond}0.850 & \cellcolor{csecond}2.437 & 0.784 & \cellcolor{csecond}0.835 & \cellcolor{csecond}2.235 \\
$(1.25,1,0.10)$ & \cellcolor{csecond}0.630 & 0.846 & 2.404 & \cellcolor{csecond}0.807 & 0.830 & 2.196 \\
\bottomrule
\end{tabular}
}
\caption{Fusion-weight sensitivity. Perturbing $(\lambda_n,\lambda_g,\lambda_o)$ around the fixed operating point $(1,1,0.10)$ leaves all metrics well above baselines, with an interpretable recall-versus-plausibility trade-off as the geometric weight $\lambda_g$ varies.}
\label{tab:lambda}
\end{table}

\setlength\belowcaptionskip{-2ex}
\begin{figure}[t]
\centering
\includegraphics[width=\columnwidth]{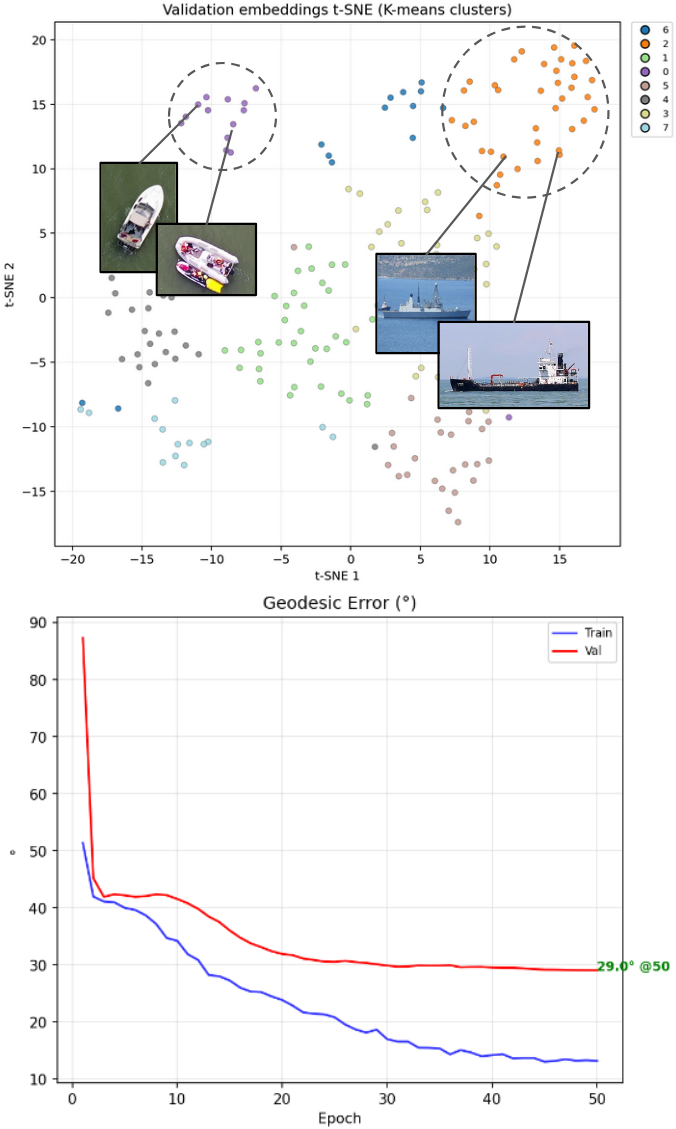}
\caption{Pose-estimator diagnostics on the validation set. \emph{Top}: t-SNE projection of the estimator's validation-crop embeddings, colored by K-means cluster; nearby crops share apparent viewpoint and orientation (e.g., near-nadir small craft versus side-view ships), so the learned features organize by pose. \emph{Bottom}: training and validation geodesic error. Both panels come from a separate $50$-epoch run on the $947$ manual annotations; Table~\ref{tab:pose_error} reports a different run.}
\label{fig:posedist}
\end{figure}
\section{Building/Training the Pose Estimator}
\label{sec:pose_supp}
\subsection{Appearance-Aligned Pose Dataset Curation}

\begin{figure}[t]
\centering
\includegraphics[width=\columnwidth]{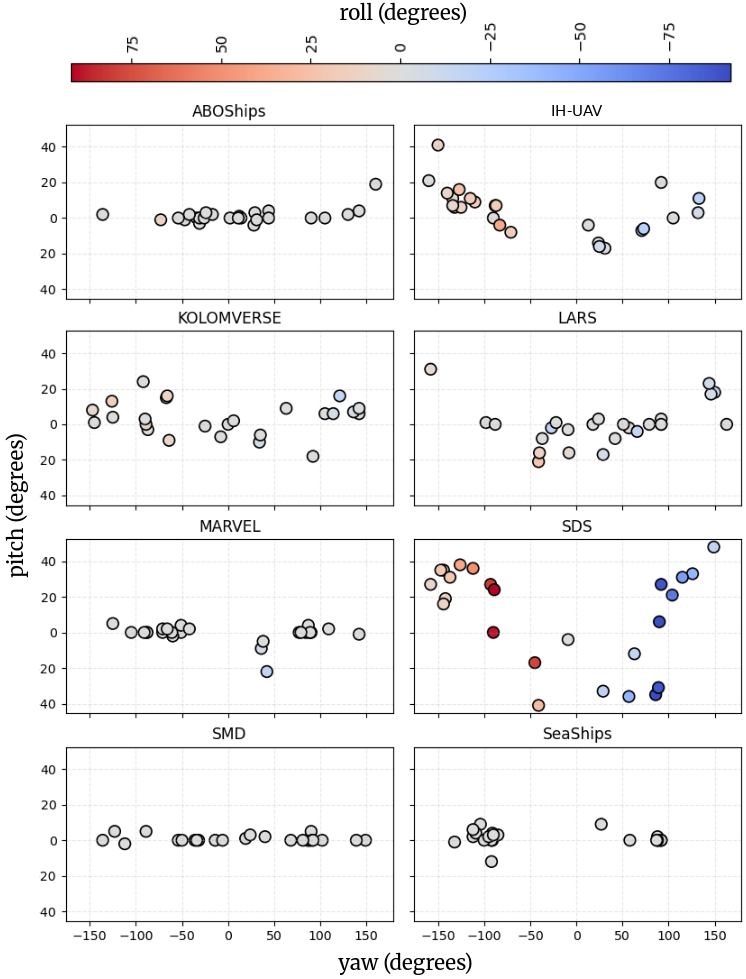}
\caption{Pose distributions across maritime datasets in our validation set. Each subplot shows vessel yaw vs. pitch, with roll indicated by color. Differences reflect varying viewpoints, e.g., aerial (IH-UAV, SDS) versus on-board/on-shore imagery (SMD, SeaShips, ABOShips, KOLOMVERSE).}
\label{valset}
\end{figure}

Pose annotations are obtained using a Unity3D interface by aligning a canonical 3D vessel model to each image instance. The reference orientation $(0,0,0)$ corresponds to a front-facing, level pose, and pitch, yaw, and roll are adjusted to match the appearance of the vessel in the image, following the \emph{apparent pose} convention. Our manual training set consists of $947$ pose annotations across \cite{varga2022seadronessee, 8438999, prasad2017video, iancu2021aboshipsinshoreoffshore, MARVEL, 10689331, zust2023larsdiversepanopticmaritime}, with a sizable portion also drawn from in-house and proprietary maritime data\footnote{Provided by the Johns Hopkins University Applied Physics Laboratory.}. Figure~\ref{fig:posedist} visualizes the trained estimator's validation embeddings and its training/validation geodesic-error convergence. In addition, our validation set is equally divided into 25 pose annotations per maritime benchmark (200 total) to ensure camera-view and vessel-type diversity (Figure~\ref{valset}).

To increase pose diversity, we augment the training dataset by applying horizontal flipping to each annotated instance. Given an original pose $(\theta_p, \theta_y, \theta_r)$, the flipped pose is computed as $(\theta_p, -\theta_y, -\theta_r)$, with angles wrapped to $[-180^\circ, 180^\circ)$, yielding an additional 900 samples from the original 900 annotations. We further incorporate $\sim$3500 samples from the BONK-Pose \cite{holst2025fusingmonocularrgbimages} dataset by converting its camera-frame rotations into our Unity-aligned frame. Given a BONK rotation $R_{\text{bonk}} \in SO(3)$, we compute pitch and roll via a fixed alignment $R_{\text{unity}} = R_{\text{bonk}} R_{\text{align}}$, while yaw is derived from the projected image-plane direction of the vessel's forward axis, yielding a pose $(\theta_p, \theta_y, \theta_r)$ consistent with our annotation convention.

\subsection{Pose Estimation}

We represent vessel orientation as a rotation matrix $R \in SO(3)$, parameterized by Euler angles $(\theta_p,\theta_y,\theta_r)$ under the image-aligned apparent pose convention. Given an input crop $I$, the estimator learns a mapping $f_\phi : I \rightarrow \hat{R}$. The model follows a dual-stream formulation, where geometric and semantic features are extracted via $F_{\mathrm{cnn}} = g_{\mathrm{cnn}}(I)$ and $F_{\mathrm{seg}} = g_{\mathrm{seg}}(I)$, respectively. The semantic branch is initialized from a SegFormer \cite{xie2021segformer} encoder trained for vessel segmentation, providing object-centric shape cues that complement geometric features. The two streams are fused as $F = \psi([F_{\mathrm{cnn}} \,\|\, F_{\mathrm{seg}}])$, and a pooled representation $\eta = \rho(F)$ is used to predict both continuous rotation and discretized pose bins.

To obtain stable predictions on $SO(3)$, we adopt the continuous 6D representation of \cite{zhou2020continuityrotationrepresentationsneural}, where a predicted vector $z \in \mathbb{R}^6$ is mapped to a valid rotation $\hat{R} \in SO(3)$. In parallel, each Euler component is discretized into $N_b$ bins via $b(\theta) = \left\lfloor \frac{\theta + \pi}{2\pi}N_b \right\rfloor$, producing categorical targets $(b_p,b_y,b_r)$ used for coarse orientation supervision and subsequent prior estimation. Training combines continuous and discrete supervision to capture both precise rotation alignment and coarse orientation structure. Specifically, we penalize (i) the geodesic distance, (ii) element-wise differences between rotation matrices to stabilize optimization, and (iii) classification error over discretized pose bins to provide coarse supervisory signals. The geodesic distance between rotations is defined as
\begin{equation}
d_{\mathrm{geo}}(R_1,R_2) = \cos^{-1}\!\left(\frac{\mathrm{Tr}(R_1^\top R_2)-1}{2}\right).
\end{equation}
The loss components are given by
\begin{equation}
\mathcal{L}_{\mathrm{geo}} = d_{\mathrm{geo}}(\hat{R}, R),
\end{equation}
\begin{equation}
\mathcal{L}_{\mathrm{6D}} = \|\hat{R} - R\|_F^2,
\end{equation}
\begin{equation}
\mathcal{L}_{\mathrm{bin}} = \frac{1}{3} \sum_{k \in \{p,y,r\}} \mathrm{CE}(\hat{b}_k, b_k),
\end{equation}
and are combined as
\begin{equation}
\mathcal{L} = \lambda_{\mathrm{geo}} \mathcal{L}_{\mathrm{geo}} + \lambda_{\mathrm{6D}} \mathcal{L}_{\mathrm{6D}} + \lambda_{\mathrm{bin}} \mathcal{L}_{\mathrm{bin}}.
\end{equation}

For vessel placement, forward and backward headings often produce equally plausible visual configurations. To account for vessel symmetry along the longitudinal axis, we also evaluate the geodesic, 6D, and yaw-bin terms under the flipped rotation $R' = R_y(\pi)R$ and keep, per sample and per term $t$, the smaller value:
\begin{equation}
\mathcal{L}_{t,\mathrm{sym}} = \min\big(\mathcal{L}_t(\hat{R}, R), \mathcal{L}_t(\hat{R}, R')\big),
\end{equation}
allowing ambiguity between forward and backward orientations.

\subsection{Pose Estimator Accuracy}
Table~\ref{tab:pose_error} reports an earlier run of this architecture, trained on the $804$ manual annotations available at the time (no flips, no BONK-Pose samples), on a validation set of $200$ vessels ($25$ per benchmark across eight datasets: the seven retrieval benchmarks plus MARVEL) with manually annotated apparent pose. The same set was used to select that run's checkpoint, and $14$ of its images also appear in the run's training annotations, so these errors are optimistic. Because vessels are longitudinally symmetric and the model is trained under the flipped-heading hypothesis ($\mathcal{L}_{\mathrm{sym}}$), we report the symmetry-aware $SO(3)$ geodesic $\min\!\big(d_{\mathrm{geo}}(\hat R,R),\,d_{\mathrm{geo}}(\hat R,R_y(\pi)R)\big)$ alongside the raw value. The estimator is accurate on the horizon-relevant axes (pitch and roll, $5.9^\circ/7.7^\circ$ MAE) and resolves yaw up to the $180^\circ$ bow/stern ambiguity (symmetric yaw MAE $18.7^\circ$). Quotienting this modeled ambiguity, the geodesic error is $23.0^\circ$ mean / $16.4^\circ$ median; the large \emph{raw} geodesic ($91.7^\circ$ mean) is almost entirely front/back heading flips (e.g., ABOShips $103.7^\circ\!\rightarrow\!13.9^\circ$ and LaRS $109.6^\circ\!\rightarrow\!28.2^\circ$ once the flip is quotiented) rather than gross orientation error. Near-nadir aerial imagery (SDS, $41.6^\circ$) remains the hardest regime. Because the scene prior weights pitch most heavily (the axis the estimator predicts most reliably), overall retrieval quality is governed by the well-estimated axes rather than the ambiguous yaw.

\begin{table}[t]
\centering
\small
\setlength{\tabcolsep}{4pt}
\resizebox{\columnwidth}{!}{
\begin{tabular}{l c cc ccc}
\toprule
& & \multicolumn{2}{c}{Geo$_{\text{sym}}\downarrow$} & \multicolumn{3}{c}{Euler MAE (deg)$\downarrow$} \\
\cmidrule(lr){3-4}\cmidrule(lr){5-7}
Dataset & $n$ & mean & med & Pitch & Yaw$^\dagger$ & Roll \\
\midrule
ABOShips   & 25 & 13.9 & 9.3  & 2.8  & 13.4 & 1.3 \\
IH-UAV     & 25 & 13.2 & 9.7  & 5.0  & 7.4  & 7.6 \\
KOLOMVERSE & 25 & 27.2 & 26.6 & 6.3  & 24.3 & 5.0 \\
LaRS       & 25 & 28.2 & 23.2 & 7.9  & 22.2 & 7.2 \\
MARVEL     & 25 & 15.4 & 10.5 & 3.8  & 12.9 & 2.7 \\
SDS        & 25 & 41.6 & 36.5 & 17.6 & 26.1 & 35.3 \\
SMD        & 25 & 29.4 & 16.8 & 1.5  & 29.0 & 1.2 \\
SeaShips   & 25 & 15.4 & 7.3  & 2.1  & 14.4 & 1.1 \\
\midrule
\textbf{Overall} & 200 & \textbf{23.0} & \textbf{16.4} & \textbf{5.9} & \textbf{18.7} & \textbf{7.7} \\
\bottomrule
\end{tabular}
}
\caption{Pose estimator accuracy on the $200$-vessel validation set (degrees), for the earlier run described in the text. Geo$_{\text{sym}}$ is the symmetry-aware $SO(3)$ geodesic that quotients the modeled $180^\circ$ heading flip (raw overall mean/median $91.7^\circ/65.8^\circ$). $^\dagger$Yaw MAE is computed under the same $180^\circ$ longitudinal symmetry. Pitch and roll (the axes the scene prior relies on most) are accurate; yaw is resolved only up to the bow/stern ambiguity.}
\label{tab:pose_error}
\end{table}

\setlength\belowcaptionskip{-2ex}

\section{BootPlace Training and Rendering}
\label{sec:bootplace}

For downstream placement evaluation, retrieved vessel candidates are converted into BootPlace-compatible insertion assets using the merged maritime COCO annotations together with SAM2-generated segmentation masks. Since the merged COCO annotations only provide bounding boxes, foreground alpha masks are generated using bbox-prompted SAM\,2~\cite{ravi2024sam2} segmentation on GPU with the \texttt{sam2\_hiera\_large} checkpoint and \texttt{sam2\_hiera\_l.yaml} configuration. Candidate crops are filtered prior to evaluation using several geometric constraints, including minimum crop size ($64{\times}64$ pixels), minimum foreground mask coverage ($5\%$), edge truncation rejection ($4$ pixel margin), and aspect-ratio filtering ($0.2 \leq w/h \leq 8.0$). Gaussian alpha feathering with radius $1.5$ is additionally applied to generated masks before RGBA export. All background scenes are resized to a fixed $960{\times}540$ evaluation resolution, while existing vessel annotations are transformed into the resized coordinate system for downstream overlap-aware filtering.

BootPlace inference uses a fixed checkpoint, trained on maritime imagery from DETR-R50 COCO weights, for all retrieval methods. BootPlace predicts the insertion center; the box size is set to that of the scene's held-out vessel. Inference uses $20$ DETR query slots, batch size $1$, $2$ dataloader workers, and CUDA execution. For Random, CLIP, DINO, DINO$+S_{\text{geom}}$, $S_{\text{geom}}$-only, and \textsc{Sealect}, predicted boxes are post-processed with IoU-based overlap rejection against existing scene vessels (threshold $0.05$). The GALA-style and DiscoFOS composites were rendered later with a variant of the composition script that skips this step but requires at least $50\%$ water beneath the hull, and they cover $210$ and $213$ of the $300$ scenes. All evaluation composites, placement metadata, overlap statistics, and visualization outputs are stored for downstream plausibility analysis. The BootPlace model is trained for $25$ epochs using the same configuration across all experiments to ensure consistent placement behavior independent of retrieval strategy.

\section{Image Harmonization}
\begin{figure}[t]
\centering
\includegraphics[width=\columnwidth]{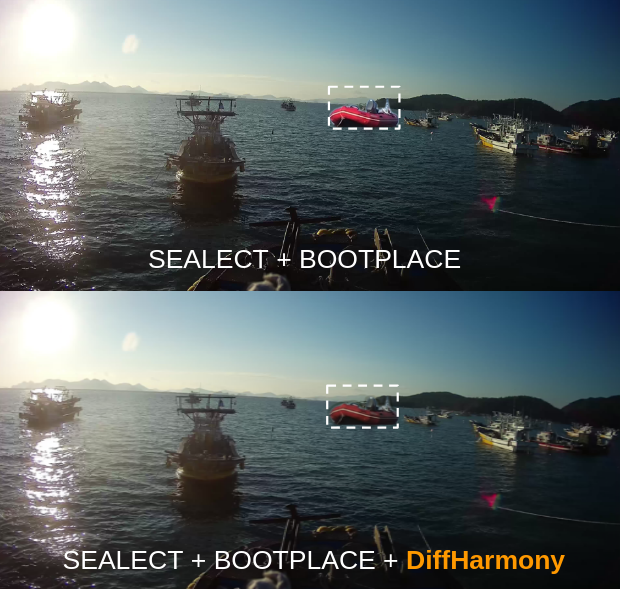}
\caption{A \textsc{Sealect} retrieval placed by BootPlace (top) and after DiffHarmony harmonization (bottom). Image source: KOLOMVERSE~\cite{10689331}.}
\label{fig:harmonization_example}
\end{figure}

\begin{figure}[t]
\centering
\includegraphics[width=\columnwidth]{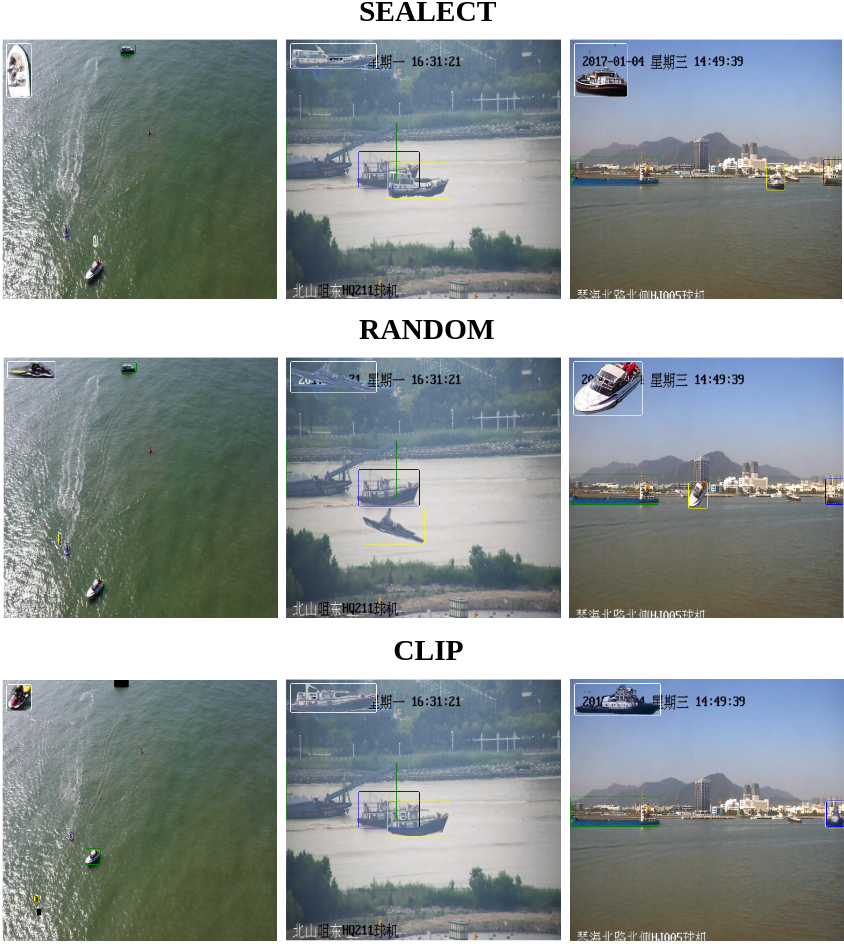}
\caption{\textsc{Sealect} retrieves vessels with plausible poses and diverse candidate structure, whereas CLIP frequently retrieves duplicate masks of vessels already present in the scene.}
\label{fig:sealect_diversity}
\end{figure}

After placement, each composite is harmonized with DiffHarmony~\cite{zhou2024diffharmony} to reconcile foreground and background color and illumination statistics. Figure~\ref{fig:harmonization_example} shows a placed-then-harmonized \textsc{Sealect} composite, and Figure~\ref{fig:sealect_diversity} contrasts \textsc{Sealect}'s pose-diverse retrievals with CLIP's tendency to duplicate vessels already present in the scene.

\section{Rendered Compatibility}
\label{sec:rendered}

\begin{table}[t]
\centering
\small
\setlength{\tabcolsep}{4pt}
\resizebox{\columnwidth}{!}{
\begin{tabular}{lcccc}
\toprule
Method & $n$ & Compat. $\uparrow$ & Mean Rank $\downarrow$ & Top-1 $\uparrow$ \\
\midrule
Random        & 300 & \cellcolor{cworst}0.285 & \cellcolor{cworst}5.68 & 6.3\%  \\
CLIP          & 300 & 0.345 & 4.64 & 11.7\% \\
DINO          & 300 & 0.370 & 4.57 & 8.0\%  \\
DiscoFOS      & 213 & 0.477 & 4.69 & 10.3\% \\
GALA-style    & 210 & \cellcolor{cbest}\textbf{0.517} & \cellcolor{csecond}3.37 & 10.5\% \\
$S_{\text{geom}}$ only$^\dagger$ & 300 & 0.429 & 4.28 & \cellcolor{cworst}3.3\%  \\
DINO$+S_{\text{geom}}$$^\dagger$ & 300 & 0.434 & 3.70 & \cellcolor{csecond}15.7\% \\
\textsc{Sealect} & 299 & \cellcolor{csecond}0.488 & \cellcolor{cbest}\textbf{3.00} & \cellcolor{cbest}\textbf{40.5}\% \\
\bottomrule
\end{tabular}
}
\caption{Rendered compatibility of each method's top-ranked candidate after BootPlace placement and DiffHarmony harmonization (score of Section~\ref{sec:eval_metrics}; $n$ scenes rendered per method). Rendered with the discrete-bin \textsc{Sealect}; $^\dagger$discrete-bin prior.}
\label{tab:rendered_pose_scene_eval}
\end{table}

We insert the top-ranked candidate of each method (top-100 pool) into $300$ test scenes with BootPlace~\cite{zhou2025bootplace} (Section~\ref{sec:bootplace}), harmonize it with DiffHarmony~\cite{zhou2024diffharmony}, and score it with the compatibility score of Section~\ref{sec:eval_metrics} (Table~\ref{tab:rendered_pose_scene_eval}). The score rates the inserted candidate's own pose, placement, and scale under the held-out reference for the context bin of the candidate's source scene; rendering affects it only through the overlap and missing-render penalties. It therefore measures candidate-level plausibility within the pipeline rather than the geometry of the final composite, and since the reference uses the kind of statistics \textsc{Sealect} optimizes, it is not prior-free; the human study (Section~\ref{sec:human}) supplies that control. GALA-style and DiscoFOS were rendered on fewer scenes with a different composition script (Section~\ref{sec:bootplace}), so ranks and top-1 rates are computed among the methods present in each scene.

\textsc{Sealect} has the best mean rank ($3.00$) and the highest top-1 rate ($40.5\%$; next best $15.7\%$). On the $210$ scenes where all eight methods are present, its top-1 rate is $39.5\%$ (next best $16.2\%$) and its mean rank is $3.21$. GALA-style's higher mean score ($0.517$ vs.\ $0.488$) comes from the penalty difference: without penalties, \textsc{Sealect} scores $0.559$ and GALA-style $0.517$. In pairwise comparisons on shared scenes, \textsc{Sealect} scores higher than GALA-style on $62.4\%$ and than DiscoFOS on $66.7\%$ of scenes.

\section{Downstream Detector Precision}
\label{sec:downstream_supp}

This experiment (Section~\ref{sec:downstream}) tests whether the geometric plausibility \textsc{Sealect} targets has a measurable effect on a downstream detector, on the precision axis. We fine-tune torchvision Faster R-CNN~\cite{ren2015faster} (ResNet-50 FPN, COCO-pretrained) for a single vessel class on a base training set of $2{,}128$ vessel-bearing SeaShips and SMD images, and evaluate cross-domain on LaRS ($2{,}605$ images, $6{,}354$ vessel boxes, $1{,}027$ vessel-empty). Within a seed, the conditions share the base image and draw the same crop and paste center from a common random generator: \emph{noaug} (no paste), \emph{plausible} (height $(0.05+0.16t)H$ growing with vertical position, upright), and \emph{implausible} (height drawn from $[0.05,0.45]H$, random vertical flip, and rotation up to $\pm35^\circ$). A paste is skipped when the vessel exceeds $90\%$ of the image or overlaps an existing vessel (IoU$>0.25$), which happens more often for the larger implausible pastes. Vessel cutouts are SAM2-masked silhouettes ($793/800$ crops) to remove paste-boundary artifacts. Each condition is trained under three seeds $\{42,123,7\}$; we report the mean$\pm$std over seeds of COCO AP, $\text{AP}_{50}$, and the mean number of detections with score $\geq0.5$ per vessel-empty image (Table~\ref{tab:downstream}); standard deviations are population values over the three seeds.

\begin{table}[h]
\centering\small
\setlength{\tabcolsep}{3pt}
\begin{tabular}{l ccc}
\toprule
Condition & AP & $\text{AP}_{50}$ & FP\,/\,empty-img \\
\midrule
No augmentation  & \cellcolor{cbest}$\mathbf{0.094}${\scriptsize$\,\pm.001$} & \cellcolor{cbest}$\mathbf{0.186}${\scriptsize$\,\pm.003$} & \cellcolor{cbest}$\mathbf{0.197}${\scriptsize$\,\pm.013$} \\
Plausible geom.  & \cellcolor{cworst}$0.086${\scriptsize$\,\pm.004$} & \cellcolor{cworst}$0.159${\scriptsize$\,\pm.005$} & \cellcolor{csecond}$0.536${\scriptsize$\,\pm.064$} \\
Implausible geom.& \cellcolor{csecond}$0.090${\scriptsize$\,\pm.002$} & \cellcolor{csecond}$0.165${\scriptsize$\,\pm.004$} & \cellcolor{cworst}$0.672${\scriptsize$\,\pm.024$} \\
\bottomrule
\end{tabular}
\caption{Downstream detector on the held-out LaRS distribution (mean$\pm$std over $3$ seeds). Faster R-CNN trained on SeaShips$+$SMD, augmented by pasting the same vessels with scene-consistent vs.\ implausible geometry. Geometric implausibility raises the out-of-distribution false-positive rate monotonically (paired in all three seeds) while AP changes little.}
\label{tab:downstream}
\end{table}

The false-positive rate is monotonic in implausibility in every paired seed (implausible $>$ plausible $>$ noaug for all three), while AP and $\text{AP}_{50}$ differ little between plausible and implausible ($\text{AP}_{50}$ is slightly higher for implausible pastes in all three seeds). The false-positive increase is consistent in sign but varies in size ($50\%$, $30\%$, and $4\%$ per seed; $25\%$ on average). Shared crops rule out candidate appearance as the cause, but the implausible condition also has larger and more often skipped pastes. This is a controlled geometry manipulation, not an end-to-end comparison of \textsc{Sealect} against appearance retrieval, and plausible augmentation does not by itself lower the false-positive rate below no augmentation. 

\section{Computational Cost and Additional Analyses}
\label{sec:extra}

\subsection{Computational Cost} Offline indexing runs one grounded-segmentation pass per library image and, per annotated vessel crop, one pose-estimator and one DINOv2 ViT-S/14 forward pass; these single-image inferences run once, after which each library entry is a compact record (scene descriptor, pose/placement/scale bins, $128$-d embedding). From these records, encoding all $68{,}338$ candidates with the candidate tower takes $3.9$\,s and assembling the kernel's library tensors $0.8$\,s on CPU. Online, scoring one query against the full pool (scene tower, neural scores, kernel prior over all $68{,}338$ library entries, observation term, per-query $z$-normalized fusion, and top-$K$) takes $11.7$\,ms median ($24.9$\,ms at the 90th percentile) on a laptop CPU (Intel i9-14900HX, 24 threads), excluding the query-side segmentation and pose passes, which cost the same per image as indexing. Scoring is exhaustive; no approximate nearest-neighbor index is needed at this scale.

\begin{table}[h]
\centering\small
\setlength{\tabcolsep}{4pt}
\begin{tabular}{l ccc c}
\toprule
$\sigma$ & ESS med. & ESS 10--90\% & ESS${<}5$ & Top-10 \\
\midrule
0.5 & 2.3 & 1.0--18.7 & 73\% & 1.00 \\
1.0 & 14.8 & 1.9--507 & 30\% & 0.68 \\
2.0 & 366 & 13.0--8{,}442 & 4\% & 0.09 \\
\bottomrule
\end{tabular}
\caption{Kernel concentration over the $68{,}338$-entry library for the $4{,}245$ test queries: effective sample size $\mathrm{ESS}=1/\sum_i w_i^2$, fraction of queries with $\mathrm{ESS}<5$, and median weight mass on the ten nearest entries. $\sigma{=}1$ is the default.}
\label{tab:ess}
\end{table}

\subsection{Kernel Concentration} Over $93$-dimensional standardized descriptors, the RBF weights of Eq.~\ref{eq:kernel_weights} do concentrate on near neighbors (Table~\ref{tab:ess}): at the default $\sigma{=}1$ the median query has an effective sample size of $14.8$ library entries, and $30\%$ of queries fall below $5$, where the prior behaves like a nearest-neighbor average. The concentration is adaptive: queries in densely populated scene regions draw on hundreds of entries ($507$ at the 90th percentile). Together with Table~\ref{tab:bandwidth}, this explains the bandwidth trade-off: sharper weights ($\sigma{=}0.5$) favor recall by matching the query's nearest scenes, while broader weights ($\sigma{=}2$, median ESS $366$) favor ranking and plausibility by averaging over more scenes.

\subsection{Observation Term} $C_{\text{obs}}$ (Eq.~\ref{eq:obs_compat}) reduces to its uniform baseline when the query scene contains no other annotated vessel, which holds for $25.6\%$ of test queries; $S_{\text{obs}}$ is then constant and, after per-query $z$-normalization, has no effect on the ranking. The remaining queries have $8.7$ other vessels in view on average. This sparsity, together with the coarse pose-bin kernel, explains the term's small contribution in Table~\ref{tab:lambda} (removing it changes Recall@20 and PS@20 by $+0.002$ and $-0.010$).

\subsection{Leakage, Camera Geometry, and Training Objective}
\noindent\textbf{Sequence-level leakage.} Leakage control (Section~\ref{sec:impl}) relies on DINOv2 similarity (near-duplicate clusters at cosine $0.95$ and image-level removal at $0.94$); without sequence labels it cannot remove consecutive frames of the same vessel that fall below these thresholds. Such pairs favor appearance-driven scores (CLIP, DINO, SigLIP\,2, and $S_{\text{neural}}$), so their absolute recall may be optimistic; $S_{\text{geom}}$ itself uses no appearance features.

\noindent\textbf{Yaw-bin concentration.} The flipped-heading yaw target in the symmetric loss (Section~\ref{sec:pose_supp}) adds $\pi$ without wrapping the angle, so for every non-negative yaw the alternative target falls in the last bin. The deployed estimator therefore predicts that bin often: across the $75{,}878$ library vessels only $9$ of the $18$ yaw bins occur, with $50.4\%$ in one bin and $32.7\%$ in another. Queries and candidates share the estimator, so yaw comparisons stay internally consistent, but yaw-dependent quantities (the yaw component of the relevance, PS, and the prior; the yaw columns of Tables~\ref{tab:dist_exp2} and~\ref{tab:plausibility_combined}; the yaw bars of Figure~\ref{fig:axis_decomp}) are coarser than the nominal $20^\circ$ bins suggest. Correcting the error requires retraining the estimator and rebuilding the library.

\noindent\textbf{Camera geometry.} Apparent pose is defined in image space and transferred between images without camera intrinsics or elevation, which our benchmarks do not provide consistently. The kernel compensates only partly, by weighting library scenes with similar viewpoint class, horizon position, and target placement, so a candidate from a scene with a very different focal length can still be scored as compatible.

\noindent\textbf{Training objective and shortcuts.} The neural retriever's in-batch InfoNCE treats every other candidate in the batch as a negative, so equally valid alternatives are sometimes penalized and the objective rewards recovering a scene's original vessel rather than its full compatible set. Our metrics grade geometric relevance rather than identity, so such alternatives still count as hits at evaluation. The scene tower receives no dataset identifier, but its scene statistics and viewpoint/layout descriptors correlate with the source benchmark, so part of the learned compatibility may reflect dataset membership rather than transferable object--scene structure.

\section{Human Plausibility Study: Full Protocol}
\label{sec:human_protocol}

\begin{figure*}[t]
\centering
\includegraphics[width=0.82\textwidth]{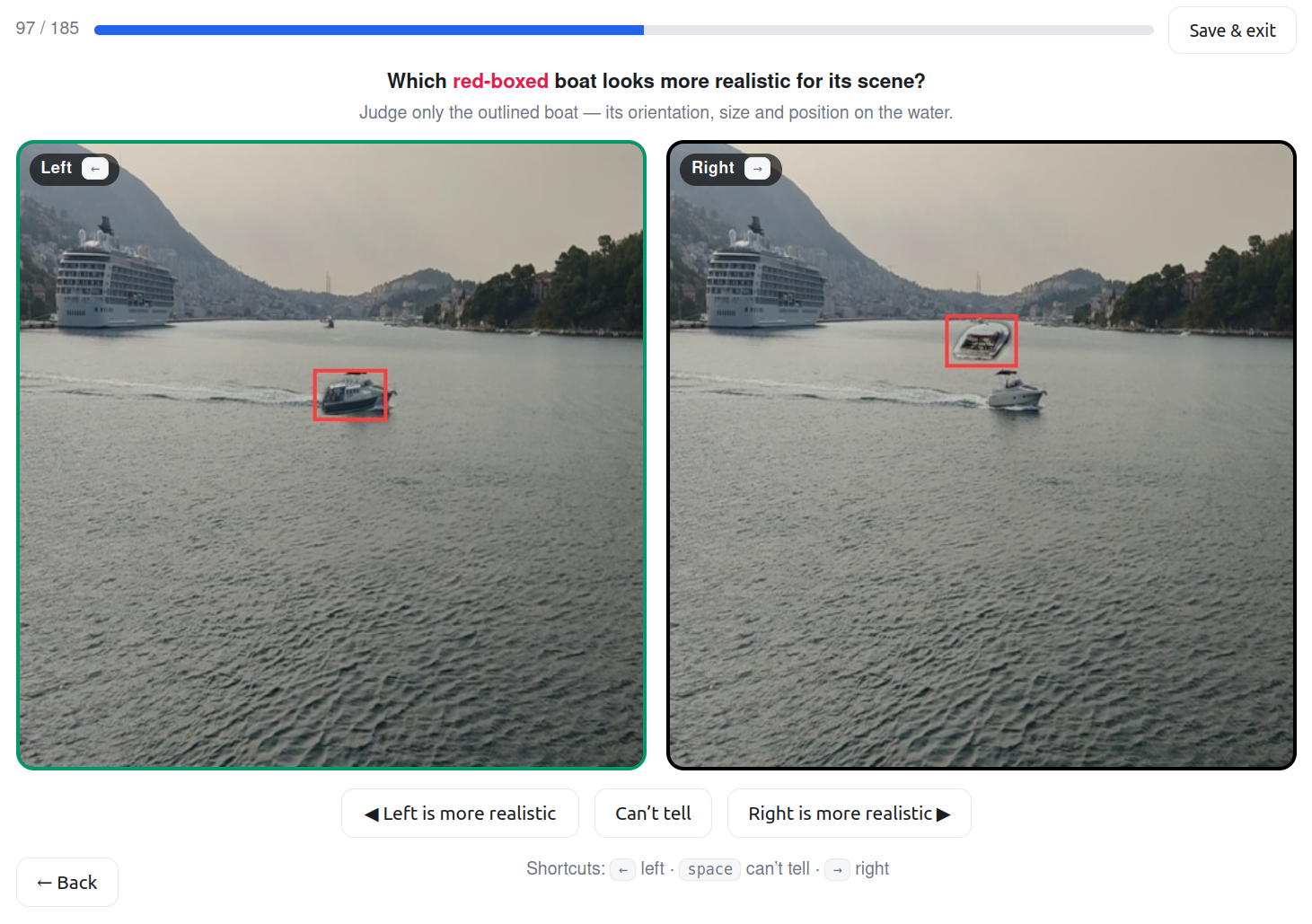}
\caption{The blind two-alternative forced-choice rating interface. Each trial presents two composites of the \emph{same} background that differ only in the inserted, red-boxed vessel, produced by two different and unlabeled methods; the rater judges only the outlined boat (its orientation, scale, and placement on the water) and selects the more realistic insertion or ``can't tell.'' Left/right assignment and trial order are randomized per rater, and no method identity is shown at any point.}
\label{fig:human_interface}
\end{figure*}

\subsection{Design} The blind study of Section~\ref{sec:human} is a paired-comparison arena with a ``can't tell'' option over the six rendered methods (\textsc{Sealect}, CLIP, DINO, DiscoFOS, GALA-style, and random selection), using the same discrete-bin \textsc{Sealect} composites as the rendered evaluation. Each trial presents two harmonized composites of the \emph{same} background that differ only in the inserted vessel, produced by two different methods; the rater selects the insertion that looks more realistic for the scene (orientation, size, and placement) or marks a tie. The schedule covers all $15$ method pairs and is weighted toward the method under test, with $25$ comparisons per \textsc{Sealect}-versus-each edge and $6$ per baseline-versus-baseline edge, so that both the primary contrasts and Bradley--Terry connectivity are well powered; this yields $185$ background-matched comparisons per rater.

\subsection{Stimulus Construction and Blinding} The inserted vessel in each composite is outlined with a colored box derived from the DiffHarmony insertion mask (rendered at $960{\times}540$ and mapped onto the $512{\times}512$ composite with an $8\%$ pad), so that raters judge only the added object rather than incidental scene differences (Fig.~\ref{fig:human_interface}). Every composite carries a neutral, method-free file name; the slot-to-method assignment is randomized by a per-comparison coin flip; and the left/right arrangement and trial order are re-randomized per rater with a seeded pseudo-random generator. No method label appears in any shipped asset, and the slot-to-method key resides only in a private manifest that is never distributed.

\subsection{Data Restriction and Artifact Control} A two-layer filter removes any comparison whose background \emph{or} inserted-vessel source is access-restricted, followed by a per-file re-audit that asserts zero restricted images; all $185$ comparisons draw from the six public datasets (ABOShips, SeaShips, SMD, KOLOMVERSE, SeaDronesSee, and LaRS). Two guards are applied when building the arena: composites containing a large solid near-black rectangle (a connected component with fill ratio $\geq0.9$) are rejected, and on non-nadir views with a confident horizon, insertions whose box bottom lies above the horizon are removed. Residual black pixels are filled by inpainting before display, and a post-hoc check confirms that dark regions are shared between the two composites of each pair. Otherwise the composites are used as the pipeline produced them, including BootPlace placements whose hull does not rest on water; only the GALA-style and DiscoFOS composites passed a water-coverage gate (Section~\ref{sec:bootplace}). Restricted to the $118$ comparisons in which both composites pass that gate, the Bradley--Terry order of the top three changes (CLIP $0.230$, DINO $0.191$, \textsc{Sealect} $0.185$), consistent with the tie reported in the main paper.

\subsection{Raters and Collection} The arena is served as a static web page that logs each response automatically, requiring nothing from raters beyond their choices. We report a primary panel of six naive external raters and a robustness panel that additionally includes the blind author, the advisor, and one partial rater who completed $68$ of $185$ comparisons, for nine raters in total; all raters are blind to method identity.

\subsection{Scoring} Method plausibility is estimated with a Bradley--Terry model~\cite{bradley1952rank} fit by minorization--maximization, with $95\%$ confidence intervals from $1{,}000$ bootstrap resamples of individual judgments; a tie counts as half a win for each side, and the head-to-head rates in the main paper exclude the $94$ ties. Resampling whole comparisons instead widens the intervals enough that \textsc{Sealect}'s overlaps random's, while GALA-style remains separated. We also compute the decisive pairwise win-rate matrix and inter-rater agreement via Fleiss' $\kappa$~\cite{fleiss1971measuring}.

\end{document}